%% file: main.tex
\documentclass[10pt,twocolumn,letterpaper]{article}

\usepackage[table,dvipsnames]{xcolor}

\usepackage{cvpr}              

\input{preamble}

\definecolor{cvprblue}{rgb}{0.21,0.49,0.74}
\usepackage[pagebackref,breaklinks,colorlinks,allcolors=cvprblue]{hyperref}

\def\paperID{*****} 
\def\confName{CVPR}
\def\confYear{2026}

\title{VGGT-SLAM++}

\author{
Avilasha Mandal$^{1}$ \quad
Rajesh Kumar$^{2}$ \quad
Sudarshan Sunil Harithas$^{3}$ \quad
Chetan Arora$^{1}$\\
\vspace{0.3em}
$^{1}$Indian Institute of Technology Delhi \quad
$^{2}$Addverb Technologies \quad
$^{3}$Brown University
}

\begin{document}
\maketitle
\input{0_abstract}

\input{1_intro}
\input{2_rel_work}

\input{3_method}

\input{4_experiment}
\input{5_discussion}

\input{7_conclusion}
{
    \small
    \bibliographystyle{ieeenat_fullname}
    \bibliography{main}
}

\input{X_suppl}

\end{document}

%% file: preamble.tex
\usepackage[margin=1in]{geometry}      
\usepackage{times}
\usepackage{epsfig}
\usepackage{graphicx}
\usepackage{float}
\usepackage{booktabs}
\usepackage{siunitx}
\usepackage{multirow}
\usepackage{array}
\usepackage{caption}
\usepackage{subcaption}
\usepackage{microtype}
\usepackage{algorithm}
\usepackage{algpseudocode} 
\usepackage{pifont}
\usepackage[utf8]{inputenc} 
\DeclareUnicodeCharacter{2009}{\,}        
\DeclareUnicodeCharacter{2248}{$\approx$} 

\definecolor{best}{RGB}{198,239,206}   
\definecolor{second}{RGB}{226,239,218} 
\definecolor{third}{RGB}{255,242,204}  
\usepackage{capt-of}
\usepackage{cuted}
\usepackage{xspace}
\usepackage{subcaption}

\newcommand{\myfirstpara}[1]{\noindent \textbf{#1.}}
\newcommand{\mypara}[1]{\vspace{0.2em} \myfirstpara{#1}}

\usepackage{amsmath,amssymb,mathtools}
\usepackage{bm}

\usepackage[font=small,labelfont=bf,tableposition=top]{caption}



%% file: 0_abstract.tex
\begin{strip}
    \centering
    \includegraphics[width=\textwidth]{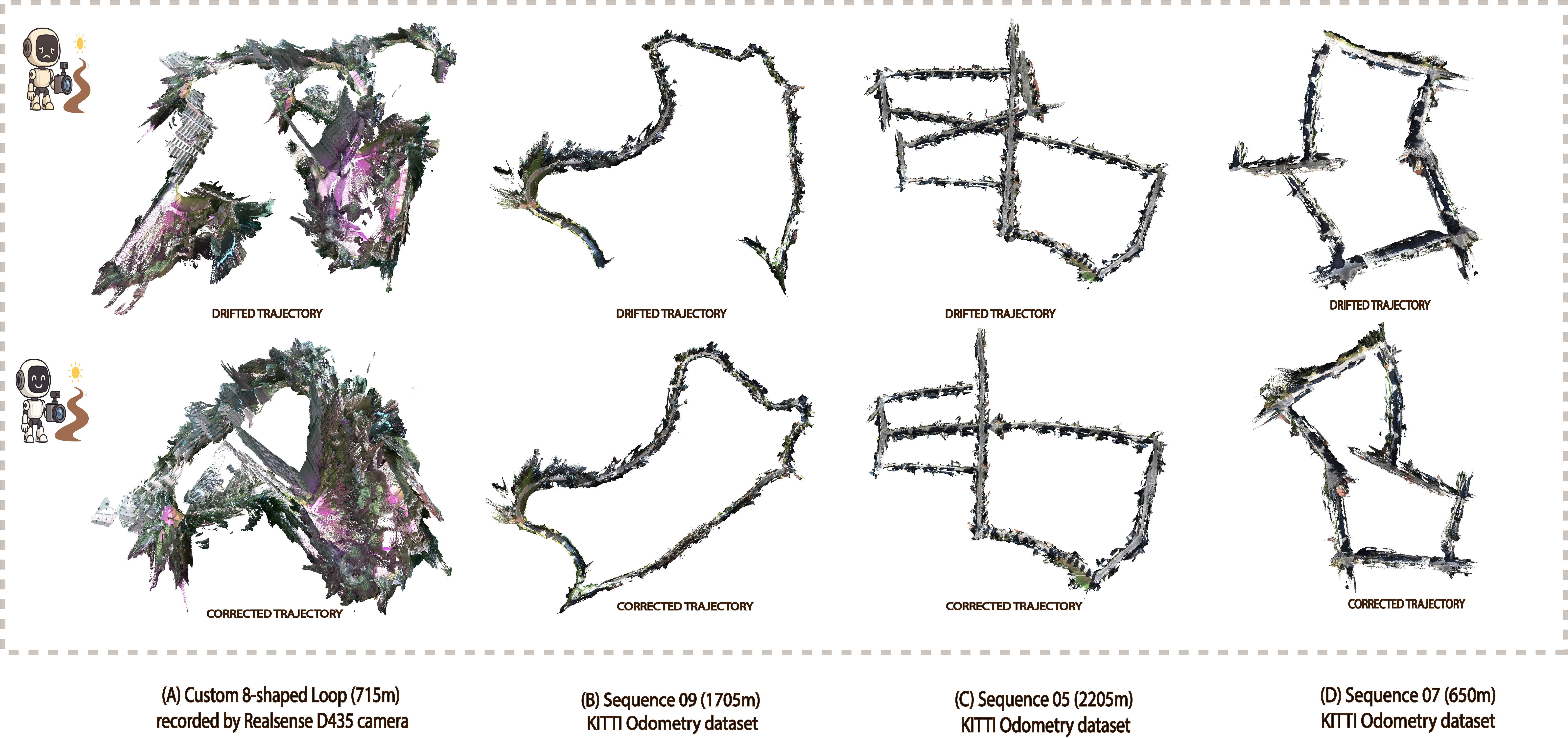}
    \captionof{figure}{ VGGT-SLAM++ provides an end-to-end SLAM architecture that stabilizes transformer-based odometry by using a low-fidelity geometric representation to support a high-cadence optimization back-end. The trajectories in the upper row are the odometry based trajectories while the lower row corresponds to the corrected trajectories when stabilised by our back-end.}
    \label{fig:teaser}
\end{strip}

\begin{abstract}
We introduce \textbf{VGGT-SLAM++}, a complete visual SLAM system that leverages the geometry-rich outputs of the Visual Geometry Grounded Transformer (VGGT). The system comprises a visual odometry (front-end) fusing the VGGT feed-forward transformer and a Sim(3) solution, a Digital Elevation Map (DEM)-based graph construction module, and a back-end that jointly enable accurate large-scale mapping with bounded memory. While prior transformer-based SLAM pipelines such as VGGT-SLAM rely primarily on sparse loop closures or global Sim(3) manifold constraints—allowing short-horizon pose drift—VGGT-SLAM++ restores high-cadence local bundle adjustment (LBA) through a spatially corrective back-end. For each VGGT submap, we construct a dense planar-canonical DEM, partition it into patches, and compute their DINOv2 embeddings to integrate the submap into a covisibility graph. Spatial neighbors are retrieved using a Visual Place Recognition (VPR) module within the covisibility window, triggering frequent local optimization that stabilizes trajectories. Across standard SLAM benchmarks, VGGT-SLAM++ achieves state-of-the-art accuracy, substantially reducing short-term drift, accelerating graph convergence, and maintaining global consistency with compact DEM tiles and sublinear retrieval. 

\end{abstract}

%% file: 1_intro.tex
\section{Introduction}
Cameras and LiDARs have been widely used for robot and autonomous vehicle localization in 3D environments for several decades \cite{thrun2002probabilistic, stachniss2016simultaneous, wang2021research, zhao2020fusion}. The camera pose estimation part of the SLAM system consists of two main components: (i) a high-frequency odometry (front-end) that provides relative pose estimates, and (ii) a slower back-end that optimizes the spatial relationships among map entities and camera poses. Classical visual odometry pipelines rely on optical-flow-based feature tracking \cite{lucas1981iterative, horn1981determining}] or feature-descriptor-based matching \cite{lowe2004distinctive,bay2006surf, rublee2011orb}, whose robustness is strongly affected by the repeatability of feature detection and the reliability of feature matching \cite{scaramuzza2011visual, sarlin2020superglue, lindenberger2023lightglue, sun2021loftr}. Recent trends emphasize the use of semantically informed \cite{mccormac2017semanticfusion, tateno2017cnn} or geometrically informed \cite{ummenhofer2017demon, teed2020raft} features and learned matchers that improve data association under challenging conditions \cite{sarlin2020superglue, detone2018superpoint}. Transformer-based architectures in computer vision has inspired a new family of SLAM and odometry systems that leverage transformer models for feature extraction, matching, and global reasoning \cite{dosovitskiy2020image, wang2025vggt, zhu2024nicer, li2023voxformer}.

Modern RGB SLAM systems \cite{zhu2024nicer, endres2012evaluation} based on dense transformers \cite{wang2025vggt} (e.g., VGGT-SLAM \cite{maggio2025vggt}) have made impressive progress in long-range consistency through robust global loop closure. However, their corrective behavior between loop events often remains coarse: the front-end \cite{nister2004visual, scaramuzza2011visual} accumulates drift, and the back-end \cite{chen2021lidar, sunderhauf2012towards} typically waits for large-baseline revisits to activate strong constraints. 
\par We address this gap with a \textbf{spatially corrective back-end} that prioritizes \textbf{high-frequency local bundle adjustment (LBA)} \cite{mur2015orb, mur2017orb, campos2021orb, sumikura2019openvslam} over exclusive reliance on infrequent global bundle adjustment (loop closures). Our key observation is that short temporal windows and neighboring submaps provide enough multi-view geometry to curb drift if they can be identified and optimized quickly. We therefore design a pipeline that (i) raises the cadence of local corrections in the back-end to keep pace with the transformer front-end, and (ii) feeds these corrections with compact, structure-preserving map evidence to make each local optimization both cheaper and more discriminative.
To this end, we introduce \textbf{DEM-augmented submaps} \cite{harithas2024findernet} and a \textbf{covisibility search} \cite{qian2022towards, mur2015orb}. Each submap exports a dense \textbf{Digital Elevation Map (DEM)}—a compact, spatially coherent projection that preserves local affine structure with enough coherent features for networks like DINOv2 \cite{zhang2022dino, oquab2023dinov2, simeoni2025dinov3} to generate structurally aware embeddings. DEMs (see Fig.~\ref{fig:DEMs}) a lightweight representation for geometry-aware alignment passes through encoders for generating candidate embeddings for retrieval. We synthesize a  covisibility graph to reduce search space and time for loop detection as we leverage, AnyLoc \cite{keetha2023anyloc} for Visual Place Recognition (VPR) within a relevant covisibilty window avoiding redundant search throughout the entire map space.
We show that \textbf{loop detection} \cite{galvez2011real, mur2015orb} via VPR \cite{lowry2015visual, keetha2023anyloc} performed directly in the DEM domain produces sufficiently informative constraints to \textbf{synthesize the spatial hierarchy}. Although DEMs provide a compact representation of large scale maps, they preserve local affine structure. It is observed that DEMs  support structural matching, we can register a new submap using a VPR module (e.g. AnyLoc  \cite{keetha2023anyloc}). Spatial hierarchy is generated within the covisibility zone of neighboring submaps. The relative pose within the  spatially connected submaps are further passed through a Sim(3) optimizer \cite{strasdat2010scale} to improve the pose of the injected submap. 

RGB images have a low field-of-view (FOV) and retrieval methods that rely on per-frame features are susceptible to false positive detections. Submap-level DEMs have much larger FOV and provide rich features.
VGGT-SLAM++ not only achieves high-cadence spatial correction of front-end drifts but also overcomes one of the key limitations of the VGGT-SLAM—in planar scenes. By leveraging a \textbf{Digital Elevation Map (DEM)}-based loop detection, our approach anchors local submaps in a geometry-preserving representation that complements transformer-derived semantics, fetching the best out of both worlds. Hence Digital Elevation Maps provide large-scale structural coherence for covisibility region discovery in long sequences (see Fig.~\ref{fig:teaser}), and precise geometric cues for local alignment and loop detection in planar environments (visualisation of trajectory in planar scene, TUM RGB-D frieburg1 floor scene is at (Fig.~\ref{fig:floor}). Consequently, our framework unifies semantic scalability and geometric stability scenario across both long-horizon and near-planar trajectories. 
The major contributions are:
\begin{itemize}
  \item Compact, geometry-preserving \textbf{DEM-augmented map representation} that remain compatible with encoders like DINOv2, enabling rich structural verification and retrieval.
  \item A \textbf{covisibility graph synthesis} via a local affine structure based search. The covisibility graph \textbf{reduces search space and complexity}, yielding faster but accurate loop detections leveraging a VPR module for candidate submap insertion within its covisibility region.
  \item A back-end that schedules \textbf{local bundle adjustment} at a higher cadence, curbing odometry drift between loop events leading to \textbf{high-frequency spatial correction.}
\end{itemize}
We validate the design on standard driving and robotics sequences, showing the presented method reduces short-term drift, accelerates graph stabilization, and maintains or improves final global consistency—while keeping runtime within practical budgets owing to the DEM-based cascaded search.

%% file: 2_rel_work.tex
\section{Related Work}




\begin{figure*}[!t]
    \centering
    \includegraphics[width=\textwidth]{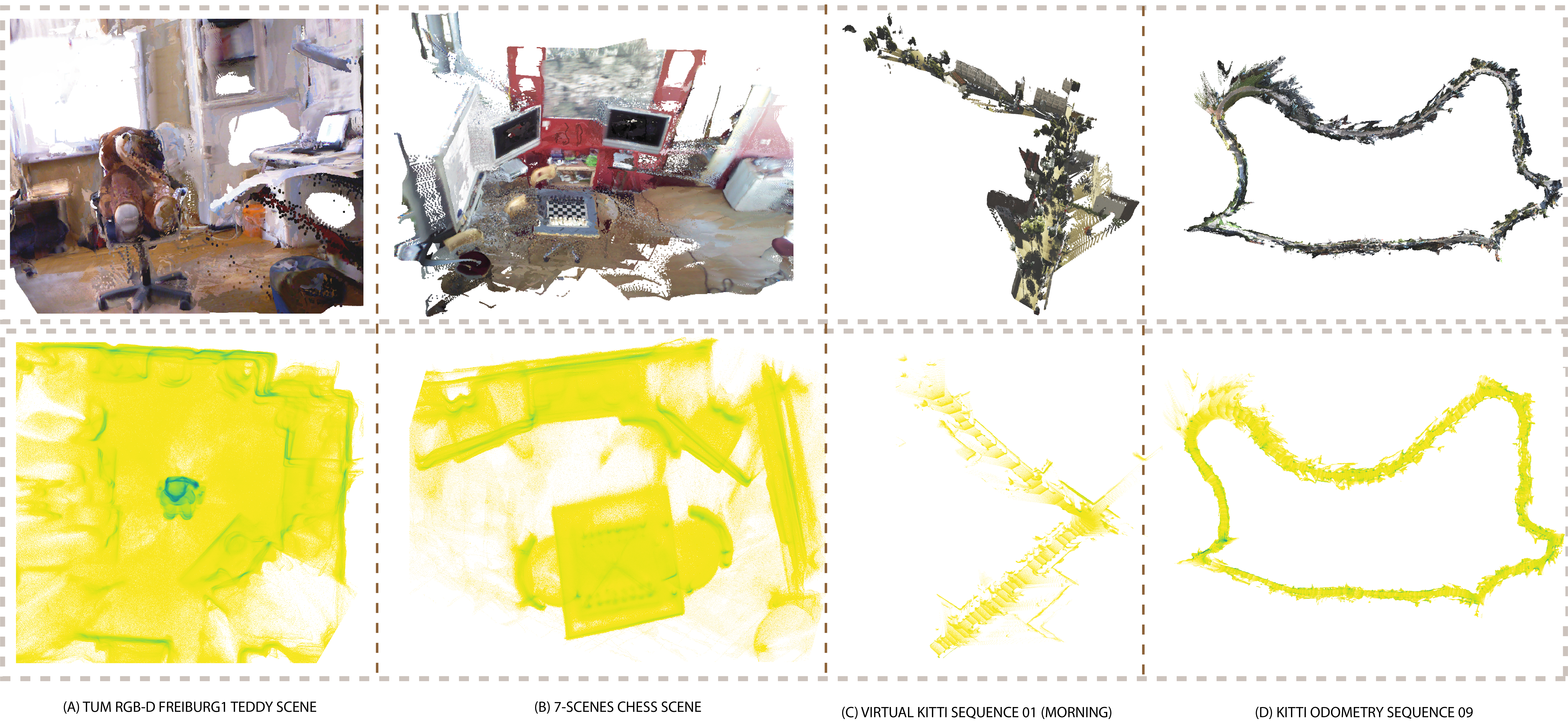}
    \caption{(A) DEM-based scene representation on the TUM RGB-D freiburg1 teddy dataset. The DEMs provide a compact 2.5D encoding retaining geometric structure.
(B) DEM visualizations from the 7-Scenes dataset.
(C) DEMs generated for the Virtual KITTI (Sequence 01) dataset.
(D) A full KITTI Odometry (Sequence 09) sequence demonstrating a complete loop, illustrating the ability of DEMs and our SLAM back-end pipeline to maintain global consistency.}
    \label{fig:DEMs}
\end{figure*}

\myfirstpara{Feed-Forward 3D Reconstruction Networks}
Feed-forward transformers have recently emerged as a unifying paradigm for multi-view 3D reconstruction, replacing iterative structure-from-motion with direct geometric inference. \textbf{DUSt3R}~\cite{wang2024dust3r} pioneered dense point and pose prediction from two uncalibrated images, followed by \textbf{MASt3R}~\cite{murai2025mast3r} and \textbf{MASt3R-SfM}~\cite{duisterhof2025mast3r}, which extended it to multi-view settings with learned correspondences and global refinement. Temporal extensions such as \textbf{Spann3R}~\cite{wang20243d} and \textbf{Cut3R}~\cite{wang2025continuous} integrate memory or recurrence for longer sequences, while \textbf{Pow3R}~\cite{jang2025pow3r} and \textbf{Splatt3R}~\cite{smart2024splatt3r} generalize the formulation to mixed cues and Gaussian-splat representations. The \textbf{MapAnything}~\cite{keetha2025mapanything} system unifies over a dozen reconstruction tasks—including multi-view stereo, SfM, registration, and depth completion—within a single feed-forward transformer that directly regresses metric 3D scene geometry. Building on this progression, the \textbf{Visual Geometry Grounded Transformer (VGGT)}~\cite{wang2025vggt} scales feed-forward reconstruction to hundreds of frames, jointly predicting cameras, depth, and dense tracks in one pass. In our work, VGGT serves as a feed-forward submap generator whose outputs provide dense geometry and camera priors for spatially corrective $\mathrm{Sim}(3)$ optimization. Following submap generation, VGGT-SLAM aligns adjacent submaps by estimating a relative transformation that resolves projective ambiguity between their respective reconstructions.



\mypara{Digital Elevation Maps (DEMs) for Compact, Structure-Aware Geometry}
DEM canonicalization \cite{harithas2024findernet} has recently emerged in LiDAR loop detection/closure \cite{chen2021overlapnet} to expose strong {planar and height priors}, achieving {large bandwidth savings} and {viewpoint robustness} via roll/pitch normalization and top-down discretization. FinderNet \cite{harithas2024findernet} demonstrates that DEMs can enable both {robust loop detection} while remaining highly {data-efficient} and {generalizable}, and that DEMs are amenable to learned embeddings without heavy augmentation.
Our use of DEMs is different in motive but similar in advantage. We generate {image-like DEMs from dense RGB submaps} (not LiDAR) as a compact {2.5D substrate} that (i) preserves {affine structure} for geometric verification, (ii) passes cleanly through {structure-aware encoders} (e.g., DINOv2 \cite{oquab2023dinov2}) for {retrieval}, and (iii) accelerates {covisibility synthesis}. This gives us compactness and geometric fidelity with structural discriminability, enabling fast and precise local corrections with reduced global complexity.

Unlike FinderNet~\cite{harithas2024findernet}, which trains a specialized network for DEM features and targets 5-DOF loop detection on dense LiDAR point clouds, VGGT-SLAM++ uses DINOv2 to embed DEMs, for transformer-generated point maps with unconstrained DOF and different noise characteristics than LiDAR. This enables robust loop detection within a complete transformer-based visual SLAM pipeline.

%% file: 3_method.tex
\section{Method}
\label{sec:method}

\begin{figure*}[!t]
    \centering
    \includegraphics[width=\textwidth]{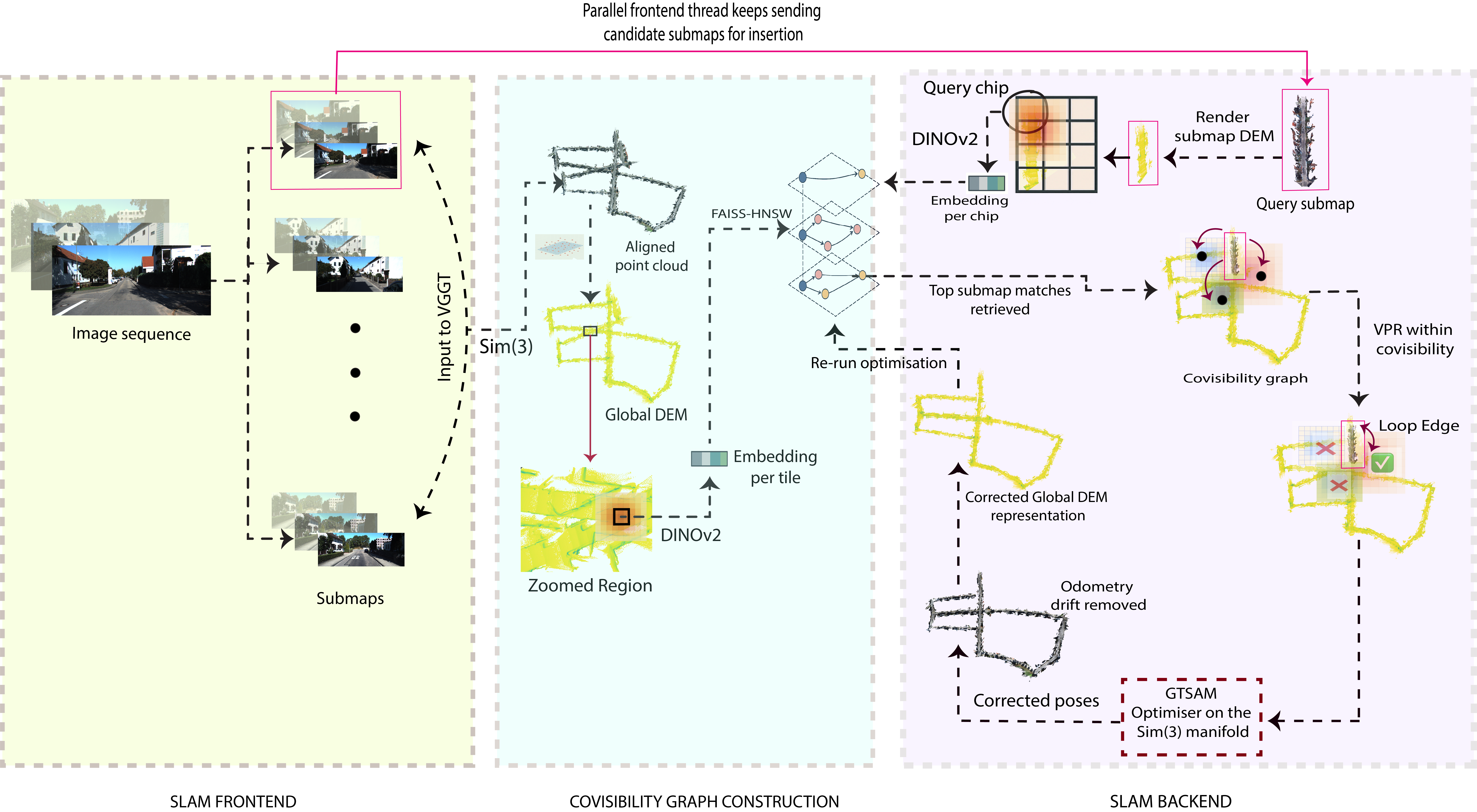}
    \caption{Complete pipeline overview.
The proposed VGGT-SLAM++ system comprises three main components:
(a) Front-end: A Sim(3) odometry module that optimizes the relative poses of submaps generated by the feed-forward VGGT network.
(b) Covisibility graph construction: A DEM-based map representation is used to compute structure-aware embeddings leveraging DINOv2, and an averaged tile score is used to insert spatially consistent nodes and edges into the covisibility graph.
(c) Back-end: An optimization module that organizes submaps into a spatial graph and performs optimization over the detected spatial constraints.}
    \label{fig:pipeline}
\end{figure*}

\mypara{Review}
VGGT-SLAM \cite{maggio2025vggt} scales VGGT \cite{wang2025vggt} to long video sequences into metrically meaningful submaps and optimizing their relative alignment. The pipeline incrementally selects keyframes from incoming RGB frames based on disparity---measured using Lucas–Kanade \cite{lucas1981iterative} flow---between the current and previous keyframes. When the disparity exceeds a threshold $\tau_{\text{disparity}}$ (we used 40m), the frame is added to the current submap’s image set $I_{\text{latest}}$. Once $|I_{\text{latest}}|$ reaches a fixed limit $w$ (we used 32), it is finalized as a submap $S_{\text{latest}}$. To ensure temporal continuity, each submap inherits one transition (non-loop closure) frame $M_{\text{prior}}$ from the previous submap, but unlike the original formulation that appended $w_{\text{loops}}$ loop-closure frames, we assign $w_{\text{loops}}=0$, to handle loop closure explicitly by our spatial drift-corrective back-end. Each submap is passed to VGGT, which reconstructs depth maps, cameras and point maps for the submap.


\mypara{Overview}
\label{subsec:overview}
Our syst
em, \textbf{VGGT-SLAM++} is an end to end SLAM framework using feed-forward transformer VGGT and a Sim(3) motion only solver. It augments the visual odometry with a spatially corrective solver over correlated submaps for continuous drift suppression. The overall pipeline (Fig.~\ref{fig:pipeline}) operates as follows: sequential RGB frames are grouped into submaps based on frame disparity and processed by VGGT to yield camera poses. Each submap is aligned to its predecessor (temporal) through a $\mathrm{Sim}(3)$ transformation, forming a temporal consistent trajectory. We do a \textbf{depth thresholding} to remove floaters from cloud or sky based noise at the reconstruction horizon. This entire (temporal) point cloud alignment of all injected submaps formed from an initial robotic tele-operation is augmented to a global planar-canonical DEM map representation which is patched into smaller tiles at 2x2 meters maintaining robust resolution. Each DEM tile is embedded into a geometric feature space using DINOv2, producing compact descriptors indexed in a FAISS-HNSW \cite{douze2025faiss} structure serving as scalable retrieval gallery, for a new query submap that arrives for registration, from the front-end tracking thread, which keeps running independently of the back-end, even after the tele-operation phase. For each such query submap, a \textbf{planar-canonical Digital Elevation Map (DEM)} is constructed in a similar fashion as the global DEM by patching into 2x2 meter chips serving as candidate queries for loop detection within a covisibility region and hence accurate registration of the submap. The front-end tracking thread continuously generates embeddings for these \textbf{query chips}—patched from candidate submaps awaiting insertion. These query embeddings of submap chips, are compared against the indexed global DEM tiles to identify spatially proximal submaps for covisibility reasoning \cite{qian2022towards, mur2015orb}. We build a covisibility window for the query submap, by comparing the DINOv2 descriptors of the 2x2 meters chips from the query submap's DEM and the 2x2 meter tiles from the global DEM. Within this covisibility window, we leverage AnyLoc \cite{qian2022towards} as a place recognition module, with the chips of submap awaiting insertion as the \textbf{queries} and the \textbf{tiles} in its covisibility region of the global DEM as the retieval gallery for each one of those queries.

\myfirstpara{DEM-Augmented Submaps}
\label{subsec:dem}
For each submap, we convert the dense 3D points into a compact, geometry-preserving \textbf{Digital Elevation Map (DEM)} defined on a single globally consistent plane. For every point obtained from VGGT's dense reconstruction, we first robustly fit a global plane using RANSAC \cite{fischler1981random} and singular value decomposition \cite{stewart1993early}. 
We compute a bounding box in the pixel domain (a tile) and choose a resolution in meters-per-pixel. The continuous pixel coordinates are then discretized into a regular grid, and all heights falling into the same pixel are aggregated via a reducer (mean, max, or softmax-weighted average, we obtain the best results with the ``softmax version''
). This yields a tiled DEM representation where each tile stores a dense 2.5D height field \cite{kagami2003vision, nam20172}.

\mypara{Structure-aware Embedding of DEM Tiles and Query Chips}
\label{subsec:embedding}
Each global DEM tile $\tau_k$ is processed through a DINOv2 \cite{oquab2023dinov2} encoder $f_\theta$ to obtain a feature vector with a weighted attention \cite{barsellotti2025talking} over a 9x9 tile neighborhood considering $\tau_k$ sitting at the center of the arrangement.
\begin{equation}v_k = 
\frac{
    \sum_j w_j\, m_j\, f_\theta(p_j)
}{
    \sum_j w_j\, m_j
},
\end{equation}
The DEM tile is first divided into small patches, producing a sequence of tokens $\{p_j\}$ that serve as the input to $f_\theta$. The scalar $w_j$ is a \textbf{Gaussian positional weight} that down-weights tokens near tile boundaries of the 9x9 neighborhood and emphasizes geometrically reliable central regions with respect to $\tau_k$. The coefficient $m_j$ is a \textbf{visibility mask} derived from the local gradient magnitude of the underlying DEM: flat or low-information regions receive lower weight, while edges, ramps, or height discontinuities contribute more strongly to the final descriptor. The resulting vector $v_k$ is therefore a normalized, geometry-aware embedding that combines structure from DINOv2 with geometric salience from the DEM.
Similar to the embedding mechanism of the global DEM tiles, every incoming query submaps are split into smaller \textbf{query chips} $\{\chi_q\}$ generated by the front-end tracking thread. These chips are passed through the same encoder $f_\theta$ with identical weighting logic (but the weighted embedding per chip is over the entire submap region patched to chips and not just a 9x9 neighborhood as in the global case) to produce query descriptors in the same embedding space. This dual embedding strategy enables new submap candidates to be compared directly against previously indexed global DEM tiles, forming the basis for fast, reliable covisibility discovery at scale.

\mypara{FAISS-HNSW Covisibility Graph Construction}
\label{subsec:covisibility}
The embedded DEM tiles populate a global FAISS-HNSW \cite{douze2025faiss} index that supports sublinear nearest-neighbor search across all previously constructed submaps. Let $v_k$ denote the DINOv2 embedding of the $k$-th DEM tile, and let $v_q$ denote the embedding of a query chip $\chi_q$ generated by the front-end tracking thread. For each query chip, we compute its similarity to every indexed DEM tile as

\begin{equation}
s(\chi_q, \tau_k)
= 
\frac{v_q^\top v_k}{\|v_q\|\,\|v_k\|},
\end{equation}

where $v_q^\top v_k$ is the dot product between the two descriptors, and $\|v_q\|$, $\|v_k\|$ are their $\ell_2$ norms. This normalized dot product corresponds to \textbf{cosine similarity} \cite{xia2015learning}, which measures structural compatibility between the chip and tile embeddings while being invariant to descriptor magnitude.

For each query chip $\{\chi_q\}$, FAISS-HNSW returns a ranked list of approximate nearest-neighbor tiles $\tau_k$ with similarity scores $s(\chi_q, \tau_k)$. We then aggregate these matches at the \emph{submap} level through simple voting: every retrieved tile contributes its raw similarity score to the score of its parent submap. Formally, the score for a submap $\mathcal{S}$ is
\begin{equation}
\mathrm{Score}(\mathcal{S})
\,{+}{=}\,
\sum_{\tau_k \in \mathcal{S}}
s(\chi_q, \tau_k),
\end{equation}
where the sum ranges over all tiles belonging to $\mathcal{S}$. Submaps whose accumulated score exceeds a similarity threshold $\tau_s$ (or rank within the top-$K$ hiararchies, we keep it to 10) are selected as \textbf{covisible neighbors} for the incoming candidate submap.

This process yields a sparse covisibility graph $\mathcal{G} = (\mathcal{V}, \mathcal{E})$, where each node in $\mathcal{V}$ corresponds to a submap, and each edge in $\mathcal{E}$ represents a strong structural relation discovered through the DEM–DINOv2 retrieval pipeline.

\mypara{Visual Place Recognition}
\label{subsec:vpr}
Once a set of top-$K$ covisible submaps has been proposed by the FAISS–HNSW voting stage, we perform a loop detection within  covisibility graph $\mathcal{G}$ using an AnyLoc-based \cite{keetha2023anyloc} Visual Place Recognition (VPR) module.
For each candidate submap pair $(i,j)\in\mathcal{E}$, AnyLoc retrieves a refined set of descriptor correspondences between the query chips and the tiles of the proposed neighbor submap. These chip-tile correspondences are once again aggregated at the \emph{submap} level through simple voting: every retrieved tile contributes its AnyLoc retieval score to the score of its parent submap. We hence obtain candidate submap-to-submap loop edges spatially from our back-end.

\mypara{Spatially Corrective Back-end Optimization}
\label{subsec:backend}
All submap-to-submap loop edges are passed to a spatially bounded back-end optimizer that operates concurrently with the front-end. Let $\{S_i\}$ denote the set of spatially connected submaps within the covisibility graph $\mathcal{G}$. For these submaps, the back-end seeks globally consistent similarity poses $\mathbf{T}_i\in\mathrm{Sim}(3)$ by minimizing the weighted geodesic error \cite{akhmedov2023using} of all loop edges:
\begin{equation}
\min_{\{\mathbf{T}_i \in \mathrm{Sim}(3)\}}
\sum_{(i,j)\in\mathcal{E}}
\Big\|
\log_{\mathrm{Sim}(3)}
\big(
\mathbf{T}_j^{-1}\,\mathbf{T}_i\,\hat{\mathbf{T}}_{ij}
\big)
\Big\|_{\Sigma_{ij}}^2.
\end{equation}
Here, $\hat{\mathbf{T}}_{ij}$ is the estimated $\mathrm{Sim}(3)$ relative transform between the submaps acting as loop edges, and $\Sigma_{ij}$ captures the per-edge uncertainty derived from descriptor consistency and 3D alignment residuals. The log-map $\log_{\mathrm{Sim}(3)}(\cdot)$ converts similarity transformations \cite{li2019robust} into their tangent-space residuals \cite{tyagi2013tangent}, enabling standard Gauss–Newton optimization \cite{burke1995gauss}.
This optimization is invoked at a high cadence and acts as a \textbf{spatially corrective layer} over the front-end: it stabilizes trajectories, suppresses drift between loop events, and maintains global consistency. Together with DEM-based geometry and transformer-derived priors, our system yields a compact, scalable SLAM system capable of robust long-horizon operation.

\begin{figure}[!t]  
    \centering
    \includegraphics[width=0.7\linewidth]{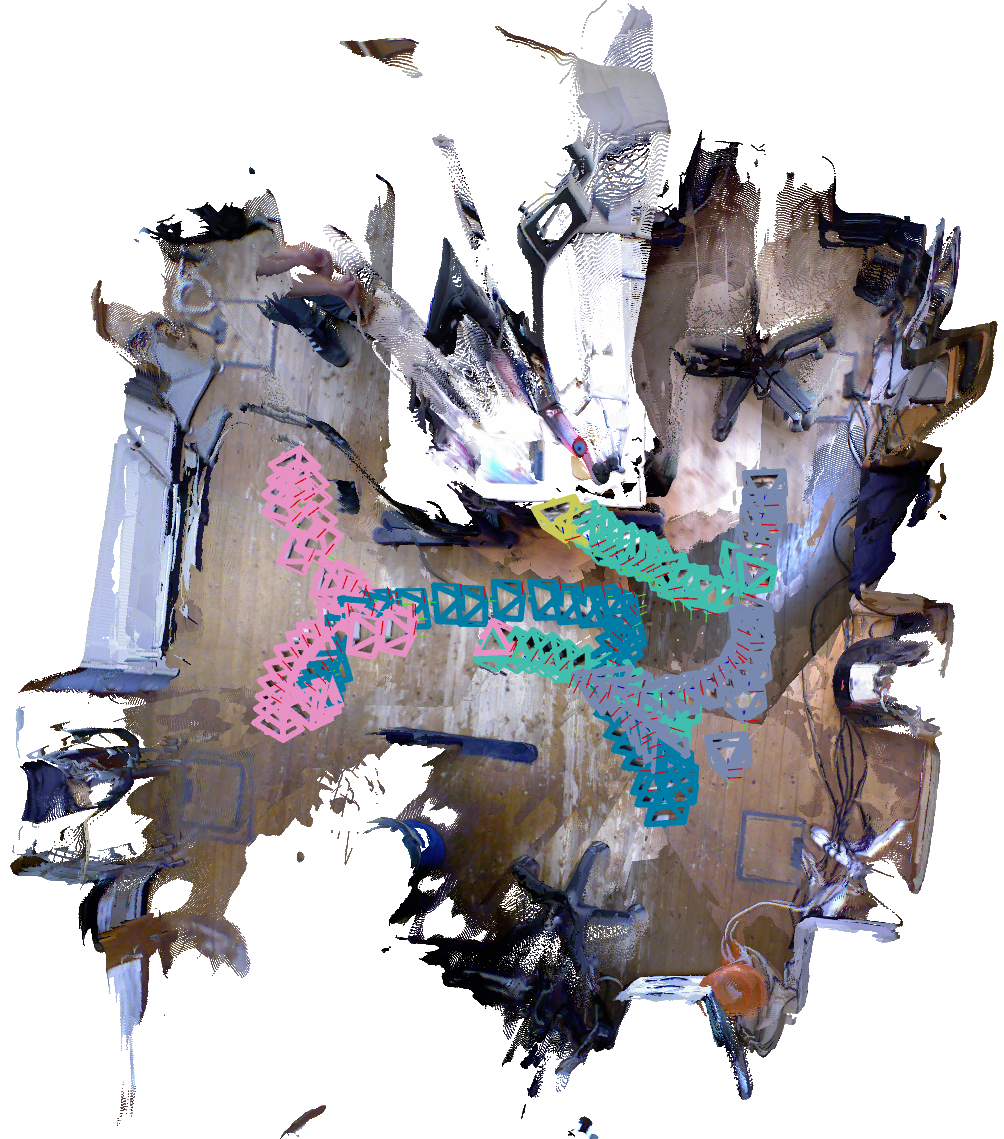}
    \caption{Preservation of geometric cues in the DEMs, alongside semantics help in accurate trajectory estimation in planar scenes like the TUM RGB-D foor scene as shown in the figure.}
    \label{fig:floor}
\end{figure}

\begin{figure}[!t]  
    \centering
    \label{reb_exp}\includegraphics[width=\linewidth]{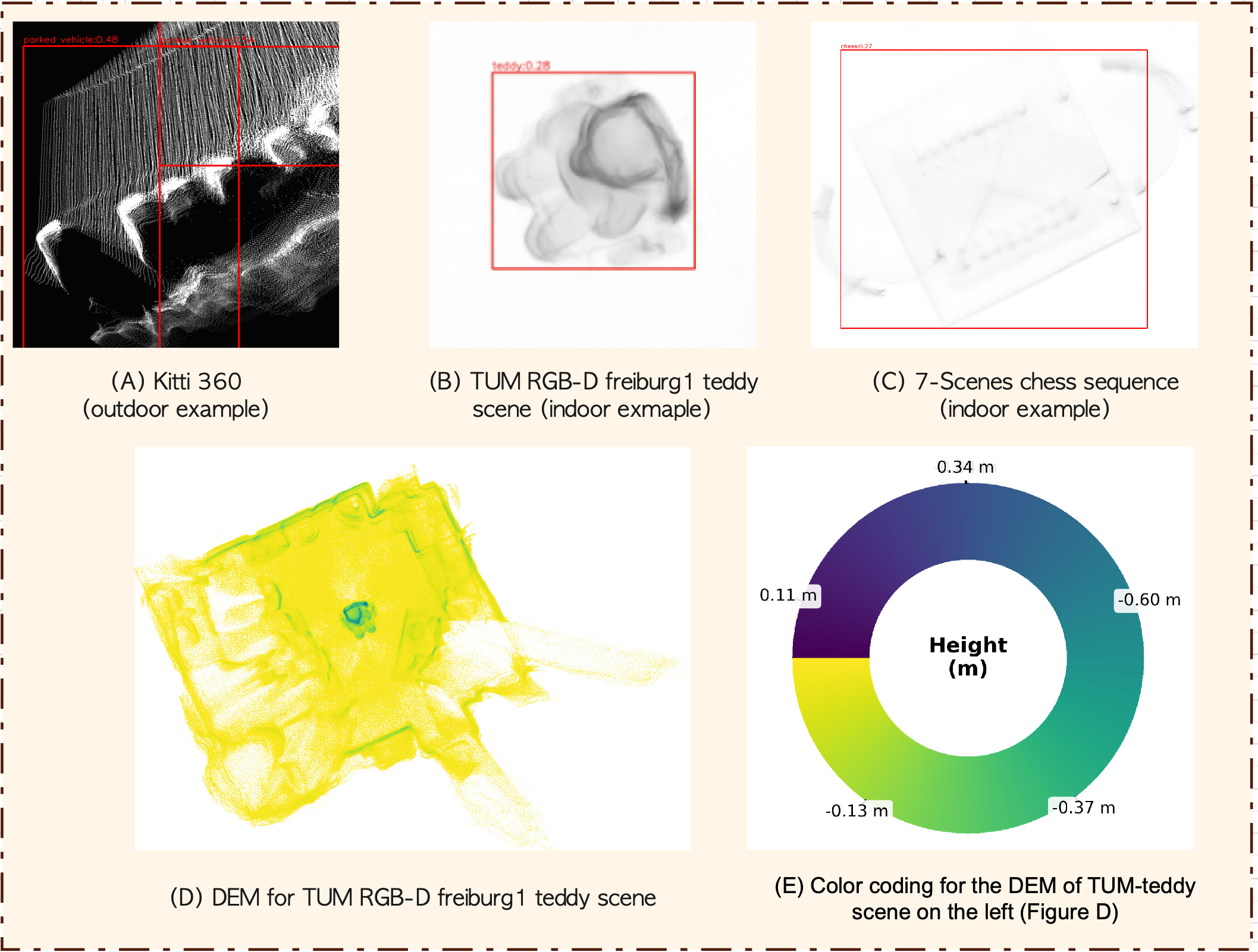}
    \caption{(A), (B), (C): zero-shot object detection from DEMs proving structure preservation. (D) DEM of TUM-teddy and (E) color coding.}
    \label{fig:proof}

\end{figure}

%% file: 4_experiment.tex
\section{Experiments}
\label{sec:experiments}

\begin{figure*}[!t]
    \centering
    \includegraphics[width=\textwidth]{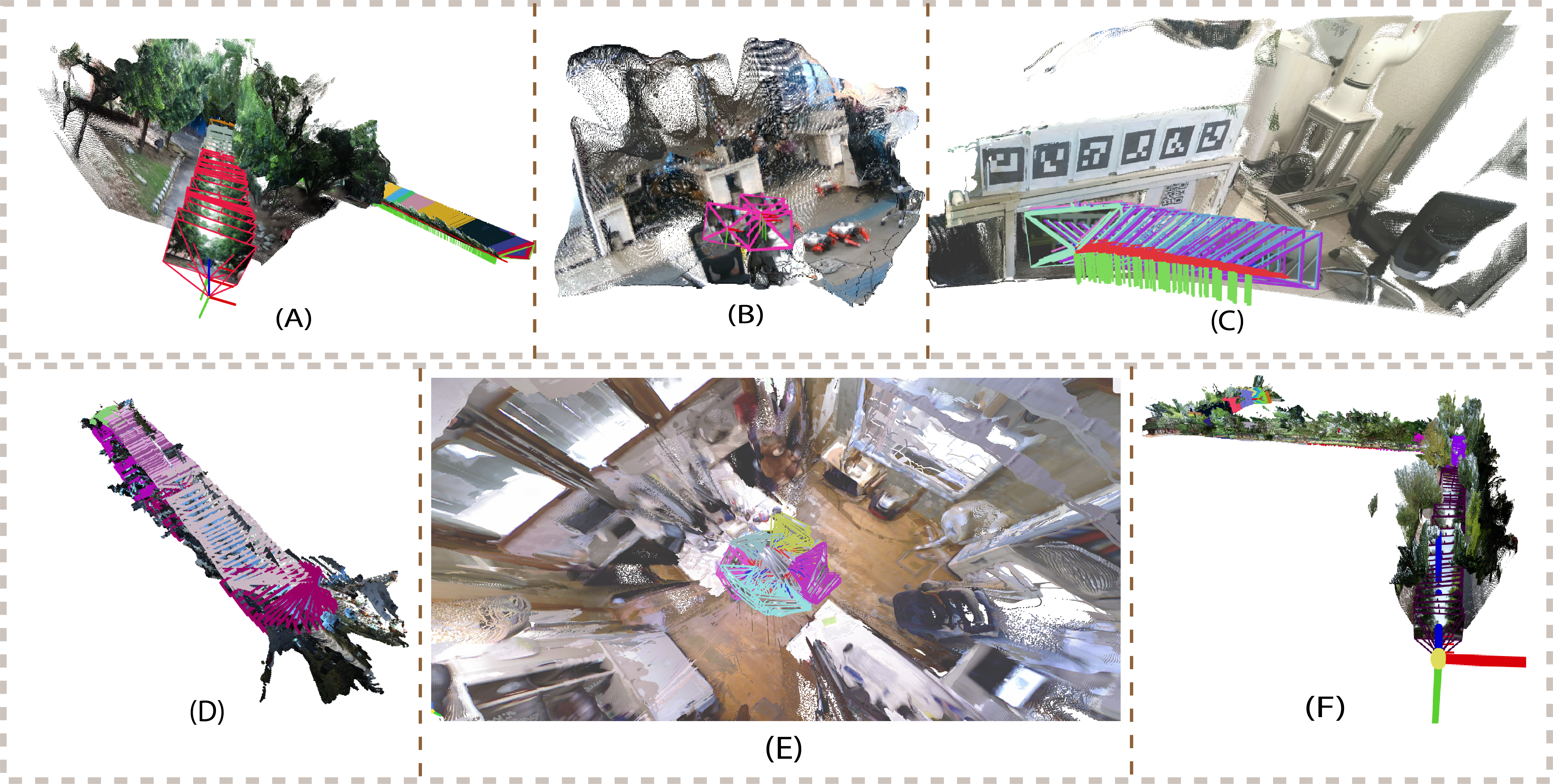}
    \caption{VGGT-SLAM++ results for :
    (A) custom data (406.8m) recorded by GoPro HERO10 camera with GPS groundtruth with 2m precision. (ATE RMSE $18 \pm 2$ m);
(B) custom data (1.8m) recorded by a OAK-1 camera with a Humanoid robot kinematics groundtruth (ATE RMSE 0.02m);
(C) custom data (1.8m) recorded by a OAK-1 camera with Cobot forward kinematics groundtruth (ATE RMSE 0.01m);
(D) KITTI Odometry 06 sequence (1230 m; ATE RMSE 13.65m);
(E)TUM RGB-D 360 scene  (5.82m; ATE RMSE 0.042m);
(F) custom data (287.381m) path recorded by the GoPro HERO10 camera with GPS groundtruth with 2m precision. (ATE RMSE $7.17 \pm 2$ m).
}
    \label{fig:results}
\end{figure*}

\myfirstpara{Datasets and Setup}
We evaluate VGGT-SLAM++ on a diverse set of datasets encompassing both synthetic and real-world conditions: 
KITTI Odometry \cite{geiger2013vision}, TUM RGB-D \cite{sturm2012benchmark}, 7-SCENES \cite{shotton2013scene}, Virtual KITTI \cite{gaidon2016virtual}, EuRoC MAV \cite{burri2016euroc} (shown in Appendix A1). Visualisations of corrected trajectories in some sequences from the above dataset are shown in Fig.~\ref{fig:results}.
All experiments are conducted on an NVIDIA RTX~4090 GPU with 24\,GB VRAM and AMD Ryzen Threadripper PRO 5955WX 16-Cores CPU with 32GB RAM. DEM rendering, DINO embedding, and local bundle adjustment (LBA) run on GPU; FAISS-HNSW indexing executes on CPU, ensuring constant memory usage per submap.
The results are benchmarked using the root mean squared Absolute Trajectory Error (ATE) \cite{zhang2018tutorial} (ATE rmse) in meters.

\mypara{Memory Profile}
At inference time, only the current submaps' (in covisibility window) VGGT features, their DEM raster, and the DINOv2 patch tokens for respective tiles from retrieval gallery or query chips reside in GPU memory [submap point clouds outside the covisibility window reside in disk], keeping VRAM usage well within the budget of an RTX\,4090 (typically 20\,GB during full operation). All global DEM tiles are stored on the CPU as compressed 2.5D grids (approximately 1--1.2\,MB each); within the 32\,GB system RAM available on the RTX 4090 server. The FAISS--HNSW index resides fully on CPU, growing sublinearly in memory due to hierarchical graph compression and fixed 768-dimensional tile descriptors. The VGGT-SLAM++ front-
end runs at $\sim$16 FPS and it's spatially corrective back-end runs at 1.89 FPS, with bounded memory usage ($\sim$8\,GB RAM, $\sim$20\,GB VRAM), averaged across datasets referred in Tables~\ref{table:kitti}, \ref{table:virtual_kitti}, \ref{table:seven_scenes}, and \ref{table:tum}, showing bounded memory compared to prior work like DROID-SLAM \cite{teed2021droid} with 8GB front-end and 24GB back-end.

\mypara{Structure-aware DEM} As shown in Fig.~\ref{fig:proof} (A, B, C) zero shot object detection by GDINO~\cite{liu2024grounding} ran on DEMs gives accurate detections for a `parked vehicle' (Kitti 360 scene), `teddy' (from TUM RGB-D freiburg1 scene) and `chess's  (7-scenes). These prove DEMs as powerful scene augmentations preserving rich structural cues.
The colored version of DEMs, shown in Figure.~\ref{fig:proof}(D, E) is only used for visualisations by humans whereas DINOv2 interpretes grayscale version. Yellow is the ground plane, darker the shade of green higher the height of the real world point.

\subsection{Quantitative Results}


Experimental results are summarized in Tables~\ref{table:kitti}--\ref{table:virtual_kitti}. As shown, \textbf{VGGT-SLAM++} achieves comparable or superior results performance across all RGB datasets with uncalibrated camera inputs, with \textbf{DROID-SLAM}, \textbf{MASt3R-SLAM}, and \textbf{VGGT-SLAM}. Notably, VGGT-SLAM++ is among the first transformer-based architectures using uncalibrated camera sources to achieve ATE comparable to calibrated-camera systems such as MASt3R-SLAM. Results on the Virtual KITTI benchmark (Table~\ref{table:virtual_kitti}) also indicate strong robustness to varying weather and illumination conditions, suggesting that the DEM representation preserves strong geometric cues. 
VGGT-SLAM++ supports both calibrated and uncalibrated versions. For completeness of the pipeline, we choose to report results with uncalibrated version, while calibrated versions are discussed in Appendix A1.

However, on uncalibrated grayscale (monochrome) datasets \cite{burri2016euroc} (results shown inAppendix A1), VGGT-SLAM++ underperforms relative to classical SLAM pipelines whose front-end odometry relies on feature tracking or optical flow rather than RGB-based transformer inference of VGGT originally trained on RGB datasets. Even in these challenging cases, the proposed method consistently improves upon VGGT-SLAM for both the \(\mathrm{Sim}(3)\) and \(\mathrm{SL}(4)\) formulations. For several long paths of the KITTI Odometry dataset \cite{geiger2013vision} monochrome sequences \cite{burri2016euroc}, the \(\mathrm{SL}(4)\) variant of VGGT-SLAM failed to converge, further underscoring the robustness and stability of the our framework.
Best is \cellcolor{best}green, 2nd is \cellcolor{second}light-green, 3rd is \cellcolor{third}yellow.
  
\begin{table}[t]
\centering
\caption{\textbf{KITTI Odometry Benchmark.} Absolute trajectory RMSE error (ATE, meters). Gray shade indicates results from classical methods. ``--'' indicates SL(4) does not converge.}
\label{table:kitti}
\resizebox{\linewidth}{!}{
\begin{tabular}{lccccccccccccc}
\toprule
\textbf{Method} & Uncalib. & 00 & 01 & 02 & 03 & 04 & 05 & 06 & 07 & 08 & 09 & 10 & Avg.\\
\midrule
\multicolumn{14}{l}{\emph{Classical feature-based SLAM}} \\
\rowcolor{gray!10}
ORB-SLAM2 (w/o LC)~\cite{mur2017orb} & \ding{55} 
& 40.65 & 502.20 & 47.82 
& \textbf{0.94} & 1.30 & 29.95 & 40.82 & 16.04 & 43.09 & 38.77 & \textbf{5.42} 
& 69.73\\

\rowcolor{gray!10}
ORB-SLAM2 (w/ LC)~\cite{mur2017orb} & \ding{55} 
& \textbf{6.03} & 508.34 & \textbf{14.76} 
& 1.02 & 1.57 & \textbf{4.04} & \textbf{11.16} & \textbf{2.19} & \textbf{38.85} & \textbf{8.39} & 6.63 
& 54.82\\

\rowcolor{gray!10}
LDSO~\cite{gao2018ldso} & \ding{55} 
& 9.32 & \textbf{11.68} & 31.98 
& 2.85 & \textbf{1.22} & 5.10 & 13.55 & 2.96 & 129.02 & 21.64 & 17.36 
& \textbf{22.43}\\
\midrule
\multicolumn{14}{l}{\emph{Learning-based SLAM}} \\

DROID-SLAM~\cite{teed2021droid} & \ding{55} 
& \cellcolor{second}92.10 & 5344.60 & \cellcolor{second}107.61 
& \cellcolor{best}2.38 & 1.00 & 118.50 & 62.47 & 21.78 & 161.60 & \cellcolor{third}72.32 & 118.70 
& 554.82\\

DPV-SLAM~\cite{lipson2024deep} & \ding{55} 
& \cellcolor{third}112.80 & \cellcolor{best}11.50 & \cellcolor{third}123.53 
& \cellcolor{second}2.50 & \cellcolor{second}0.81 & 57.80 & 54.86 & 18.77 & \cellcolor{best}110.49 & 76.66 & \cellcolor{second}13.65 
& \cellcolor{second}53.03\\

DPV-SLAM++~\cite{lipson2024deep} & \ding{55} 
& \cellcolor{best}8.30 & \cellcolor{second}11.86 & \cellcolor{best}39.64 
& \cellcolor{third}2.50 & \cellcolor{best}0.78 & \cellcolor{best}5.74 & \cellcolor{best}11.60 & \cellcolor{best}1.52 & \cellcolor{second}110.90 & 76.70 & \cellcolor{best}13.70 
& \cellcolor{best}25.75\\

VGGT-SLAM (Sim(3)) \cite{maggio2025vggt} ~\cite{orbslam3} & \ding{51} 
& 125.11 & 120.96 & 288.82 
& 5.16 & 0.96 & \cellcolor{third}29.92 & \cellcolor{third}15.03 & \cellcolor{third}14.58 & 235.80 & \cellcolor{second}38.44 & 18.60 
& 81.22\\

VGGT-SLAM (SL(4)) \cite{maggio2025vggt} ~\cite{orbslam3} & \ding{51} 
& -- & 157.01 & -- 
& 28.82 & 0.98 & -- & -- & -- & 250.72 & -- & 122.32 
& N/A\\

VGGT-SLAM++ (Ours)~\cite{orbslam3} & \ding{51} 
& 119.00 & \cellcolor{third}109.64 & 223.21 
& 4.50 & \cellcolor{third}0.95 & \cellcolor{second}25.21 & \cellcolor{second}13.65 & \cellcolor{second}12.17 & \cellcolor{third}155.00 & \cellcolor{best}35.26 & 15.71 
& \cellcolor{third}64.94\\
\bottomrule
\end{tabular}}
\end{table}

\begin{table}[t]
\centering
\caption{\textbf{TUM RGB-D Benchmark.} Absolute trajectory RMSE error (ATE, meters). Gray shade indicates results from calibrated methods. }
\label{table:tum}
\resizebox{\linewidth}{!}{
\begin{tabular}{lccccccccccc}
\toprule
\textbf{Method} & Uncalib. & 360 & desk & desk2 & floor & plant & room & rpy & teddy & xyz & \textbf{Avg.} \\
\midrule
\rowcolor{gray!10}
ORB-SLAM3 \cite{campos2021orb} & \ding{55} &
\textbf{0.017} & 0.210 & -- & \textbf{0.034} & -- & -- & -- & \textbf{0.009} & N/A & N/A \\

\rowcolor{gray!10}
DeepV2D \cite{teed2018deepv2d} & \ding{55} &
0.243 & 0.166 & 0.379 & 1.653 & 0.203 & 0.246 & 0.105 & 0.316 & 0.064 & 0.375 \\

\rowcolor{gray!10}
DeepFactors \cite{czarnowski2020deepfactors} & \ding{55} &
0.159 & 0.170 & 0.253 & 0.169 & 0.305 & 0.364 & 0.043 & 0.601 & 0.035 & 0.233 \\

\rowcolor{gray!10}
DPV-SLAM \cite{lipson2024deep} & \ding{55} &
0.112 & 0.018 & 0.029 & 0.057 & 0.021 & 0.330 & 0.030 & 0.084 & 0.010 & 0.076 \\

\rowcolor{gray!10}
DPV-SLAM++ \cite{lipson2024deep} & \ding{55} &
0.132 & 0.018 & 0.029 & 0.050 & 0.022 & \textbf{0.096} & 0.032 & 0.098 & 0.010 & 0.054 \\

\rowcolor{gray!10}
GO-SLAM \cite{zhang2023go} & \ding{55} &
0.089 & \textbf{0.016} & 0.028 & 0.025 & 0.026 & 0.052 & \textbf{0.019} & 0.048 & 0.010 & \textbf{0.035} \\

\rowcolor{gray!10}
DROID-SLAM \cite{teed2021droid} & \ding{55} &
0.111 & 0.018 & 0.042 & \textbf{0.021} & \textbf{0.016} & 0.049 & 0.026 & 0.048 & 0.012 & 0.038 \\

\rowcolor{gray!10}
MASt3R-SLAM \cite{murai2025mast3r} & \ding{55} &
0.049 & 0.016 & \textbf{0.024} &
0.025 & 0.020 & 0.061 & 0.027 &
0.041 & \textbf{0.009} & \textbf{0.030} \\
\midrule
DROID-SLAM* \cite{teed2021droid} & \ding{51} &
0.202 & \cellcolor{third}0.032 & 0.091 & \cellcolor{second}0.064 & 0.045 & 0.918 & 0.056 & 0.045 & \cellcolor{best}0.012 & 0.180 \\

MASt3R-SLAM* \cite{murai2025mast3r} & \ding{51} &
\cellcolor{second}0.070 & 0.035 & \cellcolor{third}0.055 & \cellcolor{best}0.056 & \cellcolor{third}0.035 & 0.118 & \cellcolor{third}0.041 & 0.114 & 0.020 & \cellcolor{third}0.062 \\

VGGT-SLAM (Sim3) \cite{maggio2025vggt} & \ding{51} &
0.123 & 0.040 & \cellcolor{second}0.055 & 0.254 & \cellcolor{best}0.022 & \cellcolor{second}0.088 & \cellcolor{second}0.041 & \cellcolor{second}0.032 & \cellcolor{third}0.016 & \cellcolor{second}0.079 \\

VGGT-SLAM (SL4) \cite{teed2021droid} & \ding{51} &
\cellcolor{third}0.071 & \cellcolor{best}0.025 & 0.040 & 0.141 & \cellcolor{second}0.023 & \cellcolor{third}0.102 & \cellcolor{second}0.030 & \cellcolor{third}0.034 & \cellcolor{second}0.014 & \cellcolor{second}0.053 \\

\textbf{VGGT-SLAM++ (Ours)} & \ding{51} &
\cellcolor{best}0.042 & \cellcolor{second}0.025 & \cellcolor{best}0.027 &
\cellcolor{third}0.077 & 0.042 & \cellcolor{best}0.027 & \cellcolor{best}0.026 &
\cellcolor{best}0.029 & \cellcolor{best}0.016 &
\cellcolor{best}\textbf{0.036} \\
\bottomrule
\end{tabular}}
\end{table}

\begin{table}[t]
\centering
\caption{\textbf{7-SCENES Benchmark.} Absolute trajectory RMSE error (ATE, meters). Gray shade indicates results from calibrated methods.}
\label{table:seven_scenes}
\resizebox{\linewidth}{!}{
\begin{tabular}{lccccccccc}
\toprule
\textbf{Method} & Uncalib. & chess & fire & heads & office & pumpkin & kitchen & stairs & \textbf{Avg.}\\
\midrule
\rowcolor{gray!10}
NICER-SLAM3 \cite{zhu2024nicer} & \ding{55} &
\textbf{0.033} & 0.069 & 0.042 &
0.108 & 0.200 & \textbf{0.039} & 0.108 & 0.086 \\

\rowcolor{gray!10}
DROID-SLAM \cite{teed2021droid} & \ding{55} &
0.036 & 0.027 & 0.025 &
\textbf{0.066} & 0.127 & 0.040 & 0.026 & 0.050 \\

\rowcolor{gray!10}
MASt3R-SLAM \cite{murai2025mast3r} & \ding{55} &
0.053 & \textbf{0.025} & \textbf{0.015} &
0.097 & \textbf{0.088} & 0.041 & \textbf{0.011} & \textbf{0.047} \\
\midrule

DROID-SLAM* \cite{teed2021droid} & \ding{51} &
0.047 & 0.038 & 0.034 & 0.136 & 0.166 & 0.080 & \cellcolor{second}0.044 & 0.078 \\

MASt3R-SLAM* \cite{murai2025mast3r} & \ding{51} &
0.063 & 0.046 & 0.029 & \cellcolor{best}0.103 & \cellcolor{best}0.114 & \cellcolor{third}0.074 & \cellcolor{best}0.032 & \cellcolor{second}0.066 \\

VGGT-SLAM (Sim3) \cite{maggio2025vggt} & \ding{51} &
\cellcolor{third}0.037 & \cellcolor{second}0.026 & \cellcolor{second}0.018 & 0.104 & \cellcolor{third}0.133 &
\cellcolor{second}0.061 & 0.093 & \cellcolor{third}0.067 \\

VGGT-SLAM (SL4) \cite{maggio2025vggt} & \ding{51} &
\cellcolor{second}0.036 & \cellcolor{third}0.028 & \cellcolor{third}0.018 & \cellcolor{best}0.103 & 0.133 &
\cellcolor{best}0.058 & 0.093 & \cellcolor{third}0.067 \\

\textbf{VGGT-SLAM++ (Ours)} & \ding{51} &
\cellcolor{best}0.034 & \cellcolor{best}0.023 & \cellcolor{best}0.017 &
\cellcolor{second}0.104 & \cellcolor{second}0.127 & 0.085 & \cellcolor{third}0.060 & \cellcolor{best}\textbf{0.064} \\

\bottomrule
\end{tabular}}
\end{table}

\begin{table}[t]
\centering
\caption{\textbf{Virtual KITTI Benchmark.} RMSE ATE (m). Methods shown per sequence, across weather variants.}
\label{table:virtual_kitti}
\resizebox{\linewidth}{!}{
\begin{tabular}{lcccccccc}
\toprule
\textbf{Method} & \textbf{Uncalib.} & Clone & Fog & Morning & Overcast & Rain & Sunset & \textbf{Avg.} \\
\midrule
\multicolumn{9}{c}{\textbf{Sequence 01}} \\
\midrule
DROID-SLAM \cite{teed2021droid} & \ding{55} &
\cellcolor{best}1.03 & \cellcolor{best}1.87 & \cellcolor{second}0.99 &
\cellcolor{second}1.01 & \cellcolor{best}0.78 & \cellcolor{best}1.15 &
\cellcolor{best}1.14 \\

CUT3R  \cite{khafizov2025g}    & \ding{55} &
43.30 & 63.19 & 50.60 & 38.73 & 51.55 & 43.79 & 48.53 \\

VGGT-SLAM (Sim(3)) \cite{maggio2025vggt} & \ding{51} &
\cellcolor{third}1.44 & \cellcolor{third}3.20 & \cellcolor{best}0.82 &
\cellcolor{best}0.78 & \cellcolor{second}1.56 & \cellcolor{second}3.02 &
\cellcolor{second}1.80 \\

VGGT-SLAM (SL(4)) \cite{maggio2025vggt} & \ding{51} &
3.32 & 9.02 & 1.46 & 1.74 & 5.99 & 6.21 & 4.62 \\

VGGT-SLAM++ & \ding{51} &
\cellcolor{second}1.03 & \cellcolor{second}3.06 & \cellcolor{third}1.28 &
\cellcolor{third}1.68 & \cellcolor{third}2.43 & \cellcolor{third}3.28 &
\cellcolor{third}2.13 \\
\midrule

\multicolumn{9}{c}{\textbf{Sequence 02}} \\
\midrule
DROID-SLAM \cite{teed2021droid}& \ding{55} &
\cellcolor{second}0.10 & \cellcolor{best}0.04 & \cellcolor{best}0.05 &
\cellcolor{best}0.05 & \cellcolor{best}0.04 & \cellcolor{best}0.11 &
\cellcolor{best}0.07 \\

CUT3R \cite{khafizov2025g}     & \ding{55} &
23.77 & 9.95 & 28.42 & 24.64 & 7.96 & 25.97 & 20.12 \\

VGGT-SLAM (Sim(3)) \cite{maggio2025vggt} & \ding{51} &
\cellcolor{third}0.10 & \cellcolor{second}0.15 & \cellcolor{second}0.14 &
\cellcolor{third}0.31 & \cellcolor{second}0.21 & \cellcolor{third}0.63 &
\cellcolor{third}0.26 \\

VGGT-SLAM (SL(4)) \cite{maggio2025vggt} & \ding{51} &
\cellcolor{best}0.098 & \cellcolor{third}0.15 & 0.30 &
\cellcolor{second}0.21 & \cellcolor{third}0.21 & \cellcolor{third}0.63 &
0.27 \\

VGGT-SLAM++ & \ding{51} &
0.10 & 0.15 & \cellcolor{third}0.14 & 0.31 & 0.21 & \cellcolor{second}0.60 &
\cellcolor{second}0.18 \\
\midrule

\multicolumn{9}{c}{\textbf{Sequence 06}} \\
\midrule
DROID-SLAM \cite{teed2021droid} & \ding{55} &
\cellcolor{best}0.06 & \cellcolor{best}0.02 & \cellcolor{best}0.03 &
\cellcolor{best}0.05 & TL & \cellcolor{best}0.02 &
\cellcolor{best}0.04 \\

CUT3R   \cite{khafizov2025g}   & \ding{55} &
0.84 & \cellcolor{second}0.41 & 0.60 &
\cellcolor{second}0.72 & 1.06 & 1.01 &
0.77 \\

VGGT-SLAM (Sim(3)) \cite{maggio2025vggt} & \ding{51} &
\cellcolor{second}0.10 & 0.54 & \cellcolor{third}0.14 &
0.82 & \cellcolor{best}0.28 & \cellcolor{second}0.93 &
\cellcolor{third}0.47 \\

VGGT-SLAM (SL(4)) \cite{maggio2025vggt} & \ding{51} &
\cellcolor{second}0.10 & \cellcolor{third}0.53 & \cellcolor{third}0.14 &
0.83 & \cellcolor{second}0.28 & \cellcolor{third}0.93 &
0.47 \\

VGGT-SLAM++ & \ding{51} &
\cellcolor{second}0.10 & \cellcolor{third}0.53 & \cellcolor{second}0.13 &
\cellcolor{third}0.82 & \cellcolor{third}0.28 & \cellcolor{third}0.93 &
\cellcolor{second}0.47 \\
\midrule

\multicolumn{9}{c}{\textbf{Sequence 18}} \\
\midrule
DROID-SLAM \cite{teed2021droid} & \ding{55} &
2.48 & 2.03 & 1.89 & \cellcolor{best}2.33 & 2.55 & 1.94 &
2.20 \\

CUT3R  \cite{khafizov2025g}    & \ding{55} &
19.44 & 8.63 & 6.72 & 20.21 & 16.78 & 31.12 &
17.15 \\

VGGT-SLAM (Sim(3)) \cite{maggio2025vggt} & \ding{51} &
\cellcolor{best}0.50 & \cellcolor{best}0.98 & \cellcolor{best}0.25 &
\cellcolor{third}2.57 & \cellcolor{second}2.00 & \cellcolor{best}0.36 &
\cellcolor{second}1.11 \\

VGGT-SLAM (SL(4)) \cite{maggio2025vggt} & \ding{51} &
\cellcolor{third}0.51 & \cellcolor{second}0.98 & \cellcolor{second}0.25 &
\cellcolor{third}2.57 & \cellcolor{third}2.00 & \cellcolor{second}0.36 &
\cellcolor{third}1.11 \\

VGGT-SLAM++ & \ding{51} &
\cellcolor{second}0.50 & \cellcolor{third}0.98 & \cellcolor{third}0.25 &
\cellcolor{second}2.55 & \cellcolor{best}1.99 & \cellcolor{third}0.37 &
\cellcolor{best}1.11 \\
\midrule

\multicolumn{9}{c}{\textbf{Sequence 20}} \\
\midrule
DROID-SLAM \cite{teed2021droid} & \ding{55} &
\cellcolor{third}3.59 & \cellcolor{best}5.08 & \cellcolor{best}3.73 &
\cellcolor{best}3.85 & \cellcolor{best}3.78 & 4.90 &
\cellcolor{best}4.16 \\

CUT3R  \cite{khafizov2025g}    & \ding{55} &
129.50 & 76.96 & 117.95 & 114.51 & 66.70 & 116.53 &
103.69 \\

VGGT-SLAM (Sim(3)) \cite{maggio2025vggt}  & \ding{51} &
\cellcolor{best}3.00 & \cellcolor{second}8.45 & \cellcolor{third}6.41 &
\cellcolor{second}10.00 & 6.84 & \cellcolor{best}3.64 &
\cellcolor{third}6.39 \\

VGGT-SLAM (SL(4)) \cite{maggio2025vggt} & \ding{51} &
3.87 & 9.50 & 8.21 & \cellcolor{third}10.00 &
\cellcolor{third}6.64 & \cellcolor{third}3.65 &
6.98 \\

VGGT-SLAM++ & \ding{51} &
\cellcolor{second}3.00 & \cellcolor{third}8.45 & \cellcolor{second}6.11 &
\cellcolor{second}10.00 & \cellcolor{second}5.84 & \cellcolor{second}3.64 &
\cellcolor{second}6.17 \\
\bottomrule
\end{tabular}}
\end{table}

Compared to the VGGT-SLAM baseline (Sim(3)+SL(4) averaged per dataset), VGGT-SLAM++ reduces ATE by 20\% on KITTI, 45\% on TUM, 5\% on 7-Scenes, 14\% on Virtual KITTI, 9\% on EuRoC \cite{burri2016euroc} (see Appendix A1). The combined VGGT-SLAM baseline (Sim(3)+SL(4), averaged per-dataset) results in ATE RMSE 17.13\,m whereas that of VGGT-SLAM++ is 13.94\,m, across the four datasets, hence we achive an overall improvement by \textbf{18.6\%}.

%% file: 5_discussion.tex
\subsection{Ablations}

Table~\ref{tab:kitti_dem_ablation} compares different DEM rendering choices
for VGGT-SLAM++ on KITTI odometry \cite{geiger2013vision} sequences 00–10 and reports their respective ATE RMSE (m). The best hyperparameter in every sequence is shaded in green and the second best hyperparameter is shaded light green. It has been observed that all the ablation techniques work equally well for certain sequences, hence no color shading has been done to indicate the best choice.

Let the DEM height aggregation be
\begin{equation}
\begin{aligned}
H(x,y) &= \mathrm{red}_{\tau}\!\left(\{h_{x,y}^{(k)}\}\right), \\
I(x,y) &= \mathcal{N}(H)\!
\left(1-\alpha_{\text{edge}}\|\nabla \mathcal{N}(H)\|\right).
\end{aligned}
\end{equation}
where $\mathrm{red}_{\tau}$ denotes the reducer (mean, max, or softmax temperature $\tau$), 
$\mathcal{N}(\cdot)$ is percentile normalisation, and $\alpha_{\text{edge}}$ controls Sobel-based edge shading.
Resolution is determined by the meters-per-pixel parameter $\mathrm{mpp}=S/N_{\text{px}}$.
Half/high resolution correspond to $N_{\text{px}}\!\in\!\{45k,180k\}$ (default $90k$), while
no/slight edge enhancement use $\alpha_{\text{edge}}\!\in\!\{0,0.5\}$. [Refer to Appendix A3 for more information]

\begin{table}[t]
\centering
\caption{DEM hyperparameter ablations on KITTI Odometry.}
\label{tab:kitti_dem_ablation}
\resizebox{\linewidth}{!}{
\begin{tabular}{lcccccccccccc} \toprule \textbf{Method} & 00 & 01 & 02 & 03 & 04 & 05 & 06 & 07 & 08 & 09 & 10 & Avg. \\ \midrule VGGT-SLAM++ (default) & \cellcolor{second}119.00 & 109.64 & 223.21 & 4.50 & 0.95 & 25.21 & 13.65 & 12.17 & 155.00 & 35.26 & 15.71 & 64.936 \\ 
Mean reducer (no softmax) & 120.49 & 109.64 & 223.21 & 4.50 & 0.95 & 25.21 & 13.65 & 12.17 & 155.00 & 35.26 & 15.71 & 65.072 \\ 
Softmax ($\tau=0.10$) & \cellcolor{second}119.00 & 109.64 & 223.21 & 4.50 & 0.95 & 25.21 & 13.65 & 12.17 & 155.00 & 35.26 & 15.71 & 64.936 \\ 
Softmax ($\tau=0.005$) & 120.49 & 109.64 & \cellcolor{second}219.25 & 4.50 & 0.95 & 25.21 & 13.65 & 12.17 & 155.00 & 35.26 & 15.71 & \cellcolor{second}64.711 \\ 
Half resolution & \cellcolor{best}116.53 & 109.64 & \cellcolor{best}171.42 & 4.50 & 0.95 & \cellcolor{best}20.79 & \cellcolor{best}5.85 & 12.17 & 155.00 & 35.26 & 15.71 & \cellcolor{best}58.893 \\ 
High resolution & 132.54 & 109.64 & 223.21 & 4.50 & 0.95 & \cellcolor{second}23.32 & 13.65 & 12.17 & 155.00 & 35.26 & 15.71 & 65.995 \\ 
No edge-enhancement & 120.49 & 109.64 & \cellcolor{second}219.25 & 4.50 & 0.95 & 25.21 & 13.65 & 12.17 & 155.00 & 35.26 & 15.71 & \cellcolor{second}64.711 \\ 
Slight edge-enhancement & \cellcolor{second}119.00 & 109.64 & 223.21 & 4.50 & 0.95 & 25.21 & 13.65 & 12.17 & 155.00 & 35.26 & 15.71 & 64.936 \\

\end{tabular}}
\end{table}

\subsection{Discussion}
Our method shows consistent improvements in both ATE and runtime efficiency while maintaining low memory usage.
VGGT-SLAM++ achieves near real-time operation, confirming the benefits of spatially corrective optimization and DEM-based compactness. 
\par An observation is the limited ability of the VGGT odometry module to provide accurate motion estimates on monochrome (grayscale) datasets such as \textbf{EuRoC} \cite{burri2016euroc} (Tables and discussion in Appendix A1), as the underlying transformer was trained exclusively on RGB data. However, the proposed back-end framework significantly improves the Absolute Trajectory Error (ATE) by \textbf{18.6\%} (across all five datasets), relative to the VGGT-SLAM baseline.

%% file: 7_conclusion.tex
\section{Conclusion}
We presented \textbf{VGGT-SLAM++}, a transformer-based visual SLAM system that couples VGGT-derived odometry with a DEM-based covisibility framework and a local bundle adjustment. By representing each submap as a compact DEM, the system preserves essential structural cues of a scene while enabling efficient retrieval through DINOv2 embeddings. We achieve SOTA accuracy on RGB datasets and delivers notable improvements on monochrome sequences where feed-forward transformer odometry is less reliable. The DEM representation also provides accuracy gains with minimal computational cost, making the system well-suited for real-time deployment on edge platforms. Future work will explore model compression and multi-modal sensing to further improve computational burden, and generalization.

\mypara{Acknowledgement}
We thank Aryan Singh for assistance
with some of the experiments. This research was conducted in collaboration with Addverb Technologies and IHFC.

%% file: X_suppl.tex
\section*{Appendix}

Appendix provides additional mathematical and algorithmic details that
underpin the design of VGGT-SLAM++.  We focus on six components: 
(A1) VGGT-SLAM++ Result Discussion on Established and Custom Datasets
(A2) Depth Thresholding and Removal of Far-Field Floaters
(A3) Global DEM Construction and Colour Mapping,  
(A4) FAISS--HNSW as the Covisibility Retrieval Backbone,  
(A5) Use of AnyLoc on DEM Images,  
(A6) Choice of $\mathrm{Sim}(3)$ for Back-end Optimisation

\subsection*{A1.\quad VGGT-SLAM++ Result Discussion on Established and Custom Datasets}

We have performed the experiments with VGGT-SLAM++ over 5 established datasets, KITTI odometry \cite{geiger2013vision}, Virtual KITTI \cite{gaidon2016virtual}, TUM RGB-D \cite{sturm2012benchmark}, 7-Scenes \cite{shotton2013scene}, EuRoC-MAV \cite{burri2016euroc}. Our corrective backend leads to drift correction with loop detections and closures.

An observation is the limited ability of the VGGT odometry module to provide accurate motion estimates on monochrome (grayscale) datasets such as \textbf{EuRoC} (shown in table~\ref{table:euroc_mav}; best result is \cellcolor{best}green, second best is \cellcolor{second}light-green, third best is \cellcolor{third}yellow.), as the underlying transformer was trained exclusively on RGB data, hence we observe under-performance of VGGT-SLAM++ compared to classical methods and DROID SLAM~\cite{teed2021droid}. Yet VGGT-SLAM++ reduces ATE by 9\% on EuRoC compared to the VGGT-SLAM baseline (Sim(3)+SL(4) averaged), due to our drift corrector backend module.

\begin{figure}[!t]  
    \centering
    \includegraphics[width=0.7\linewidth]{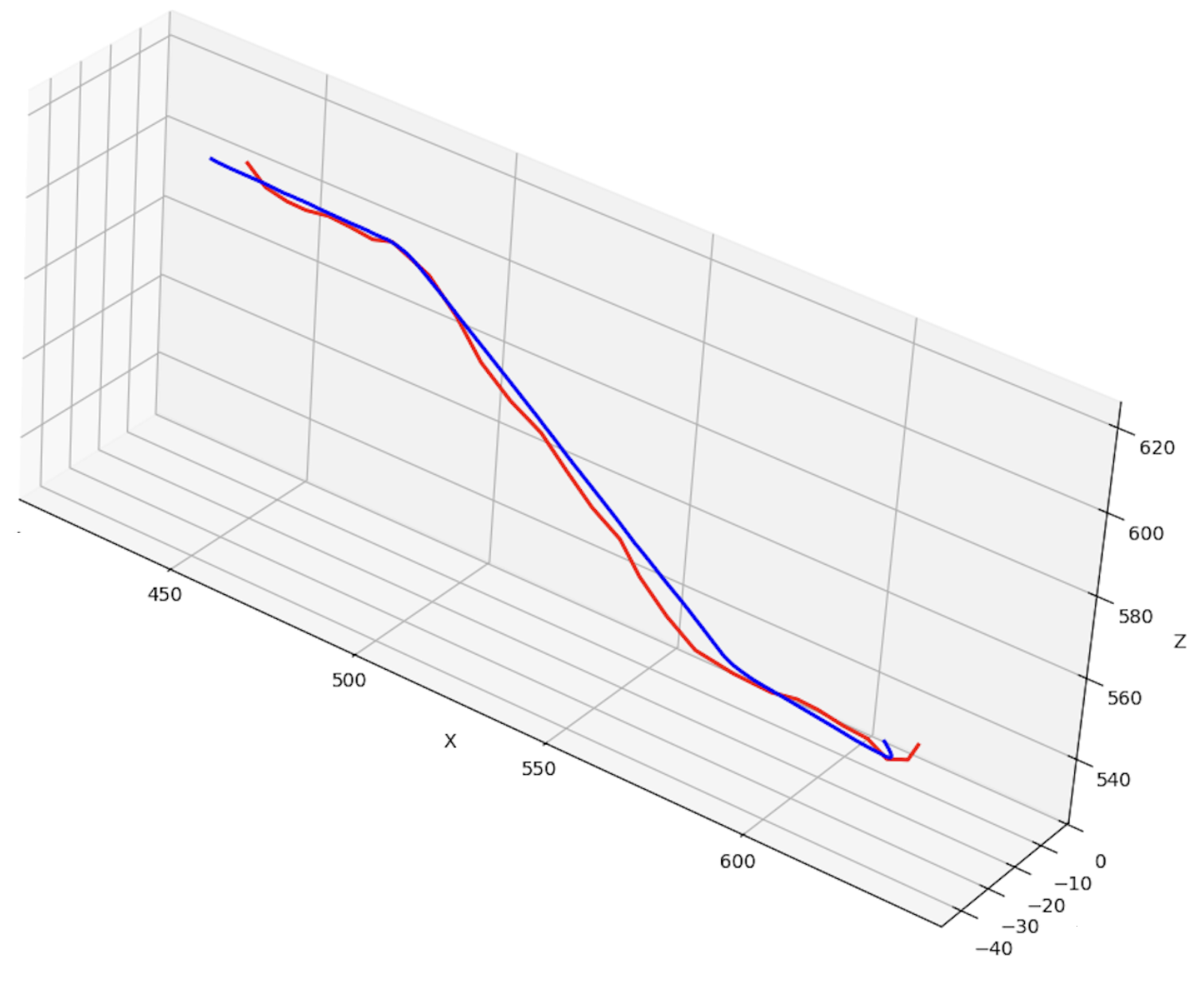}
    \caption{The red line is the ground truth reference from GPS readings and the blue line is the estimation by VGGT-SLAM++ for the custom GoPro camera dataset [Axes' units are in meters].}
    \label{fig:custom}
\end{figure}

We also extend our experiments over custom datasets with various ground truth sources, such as the custom
dataset (Fig 5(F) from the main paper) with 287.381m long path length recording by the GoPro HERO10 camera with GPS groundtruth (with precision 2m) from the Geo Tracker mobile application (ATE RMSE 7.17 $\pm 2$ m) as shown in Fig. \ref{fig:custom}. We have also conducted several experiments with custom recordings by the OAK-1 camera (Fig 5(B) from the main paper) with Humanoid robot kinematics groundtruth (ATE RMSE
0.02m) over path length of 1.8m and another case, also with the OAK-1 based camera recording (Fig 5(C) from the main paper) with Cobot forward kinematics groundtruth (ATE RMSE 0.01m) over path length of 1.8m (a planar scene, showing the ability of DEMs to make the trajectory estimation accurate even in planar domain). We also show that the (Digital Elevation Maps) DEMs \cite{harithas2024findernet} can handle the loop detection while re-localising a place, even from opposite ends with a completely different front view as they are inherently based upon the top view geometry which is constant while approaching the place from either sides. The center point of the 8-shaped loop ((Fig 1(A) from the main paper)) is reached from opposite ends leading to different front-views but since DEMs are agnostic to this fact, with their property of rendering the canonical height map of the place, we can detect loops in a viewpoint invariant style.

VGGT-SLAM++ supports both calibrated and uncalibrated versions. On KITTI, known intrinsics yields marginal gains: Seq.~05 improves from 25.21\,m to 25.20\,m; Seq.~03 remains at 4.50\,m. Our novelty lies in engineering a backend, which cuts drift at high cadence, agnostic to the fact of whether calibration exists or not.

\begin{table}[t]
\centering
\caption{\textbf{EuRoC MAV Benchmark.} Absolute trajectory RMSE error (ATE, meters). Gray shade indicates results from calibrated methods. ``--'' indicates SL(4) does not converge.}
\label{table:euroc_mav}
\resizebox{\linewidth}{!}{
\begin{tabular}{lccccccc}
\toprule
\textbf{Method} & Uncalib. & MH01 & MH02 & MH03 & MH04 & MH05 & \textbf{Avg.}\\
\midrule
\rowcolor{gray!10}
ORB-SLAM~\cite{mur2015orb} & \ding{55} &
0.071 & 0.067 & 0.071 & 0.082 & \textbf{0.060} & 0.070 \\

\rowcolor{gray!10}
DSM~\cite{zubizarreta2020direct} & \ding{55} &
0.039 & 0.036 & 0.055 & \textbf{0.057} & 0.067 & \textbf{0.051} \\

\rowcolor{gray!10}
ORB-SLAM3~\cite{mur2015orb} & \ding{55} &
\textbf{0.016} & \textbf{0.027} & \textbf{0.028} & 0.138 & 0.072 & 0.056 \\
\midrule
DeepFactors~\cite{czarnowski2020deepfactors} & \ding{51} &
\cellcolor{second}1.587 & \cellcolor{second}1.479 & 3.139 & 5.331 & \cellcolor{second}4.002 & 3.108 \\

DROID-SLAM~\cite{teed2021droid} & \ding{51} &
\cellcolor{best}0.013 & \cellcolor{best}0.014 & \cellcolor{best}0.022 &
\cellcolor{best}0.043 & \cellcolor{best}0.043 & \cellcolor{best}0.027 \\

VGGT-SLAM (Sim(3))~\cite{maggio2025vggt} & \ding{51} &
1.740 & 2.890 & \cellcolor{third}2.270 &
\cellcolor{third}3.390 & 4.400 & \cellcolor{third}2.938 \\

VGGT-SLAM (SL(4))~\cite{maggio2025vggt} & \ding{51} &
3.780 & 3.960 & 3.710 & -- & -- & N/A \\

VGGT-SLAM++ (Ours) & \ding{51} &
\cellcolor{third}1.600 & \cellcolor{third}2.700 & \cellcolor{second}1.900 &
\cellcolor{second}2.980 & \cellcolor{third}4.150 & \cellcolor{second}2.666 \\
\bottomrule
\end{tabular}}
\end{table}

\subsection*{A2.\quad Depth Thresholding and Removal of Far-Field Floaters}

A recurrent failure mode in dense transformer reconstructions is the presence of
\emph{far-field floaters} \cite{wirth2023post, yang2024clear}: points reconstructed at extremely large depth due to
 textureless sky, horizon regions, or ambiguous background surfaces.  These
points do not correspond to observable geometry and, if left unfiltered,
produce high-elevation spikes in the DEM that violate the planar assumption and might
introduce unstable gradients for both DINOv2 \cite{oquab2023dinov2} embeddings and the Sim(3) backend \cite{strasdat2010scale}.
To prevent this, VGGT--SLAM++ applies a physically-motivated depth filter
\[
  \forall p_i = (x_i,y_i,z_i)^\top \in P
\]
\begin{equation}
    p_i \;\text{``is kept'' if}\;
  d_{\min} \le \|p_i\|_2 \le d_{\max}, 
\end{equation}
where $d_{\min}$ and $d_{\max}$ are user-specified bounds that remove implausibly near or implausibly distant structures.  In practice, points with
$\|p_i\|_2 \gg 30\,$m typically originate from ambiguous sky pixels or regions
with vanishing disparity; these inflate the DEM by acting as outlying ``mountain
peaks'' during softmax aggregation \cite{wang2023improved}.  Filtering them ensures that the retained
set
\begin{equation}
  P_{\mathrm{valid}}
  = \{\, p_i \in P : d_{\min} \le \|p_i\|_2 \le d_{\max} \,\}
\end{equation}
spans genuine scene geometry.  This stabilises subsequent steps: (i)
plane-fitting \cite{fischler1981random} becomes robust because extreme outliers no longer dominate the
covariance; (ii) height aggregation behaves smoothly because all samples within
a pixel correspond to metrically reasonable depths; and (iii) global DEM tiles
exhibit clean, horizon-free elevation fields without sky-induced artefacts.
Depth thresholding therefore plays the same role for height stability as
confidence filtering does for prediction quality, ensuring that VGGT--SLAM++
builds DEMs solely from geometrically meaningful 3D structure.

\begin{figure}[!t]  
    \centering
    \includegraphics[width=0.7\linewidth]{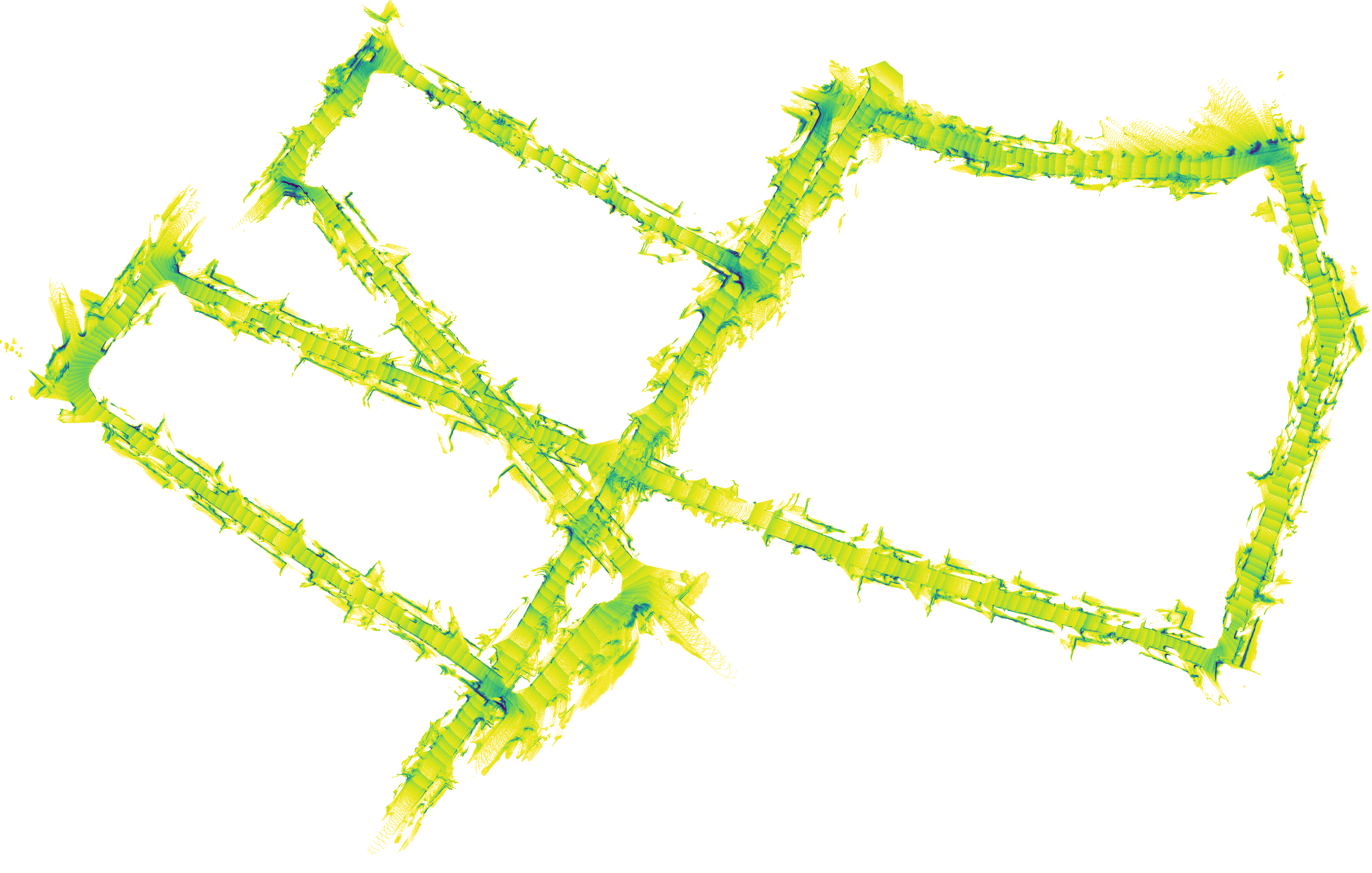}
    \caption{The DEM rendered from the 3D points aligned by odometry over  the KITTI Sequence 05, with color mapping for better visualisation.}
    \label{fig:dem}
\end{figure}

\subsection*{A3.\quad Global DEM Construction and Colour Mapping}

This section explains construction of a
global Digital Elevation Map (DEM) \cite{harithas2024findernet} from  3D points generated by a feedforward transformer \cite{wang2025vggt}.  The goal is to construct a planar-canonical DEM whose domain is a large rectangular region in a dominant
ground-like plane and whose values encode signed height above that plane.  The
process consists of (i) fitting a stable reference plane, (ii) expressing all
points in a canonical orthonormal frame, (iii) rasterising heights at a chosen
spatial resolution, and (iv) feeding the grayscale DEM for DINOv2-based retrieval.

\paragraph{Input.}
Let 
\begin{equation}
  P = \{p_i\}_{i=1}^N, 
  \qquad p_i = (x_i,y_i,z_i)^\top\in\mathbb{R}^3,
\end{equation}
denote all 3D points reconstructed by the frontend in a common world frame.
These points arise from VGGT depth and Sim(3) odometry.
\paragraph{1.\ Plane fitting.}
Inherently a dominant structure underlies every practical scene from a standard dataset e.g.\ the ground
plane, floor, or road is approximately planar over the global scale (comprising maximum number of points).  We fit
a plane
\begin{equation}
  \Pi = \{p\in\mathbb{R}^3 : n^\top p + d = 0\},
\end{equation}
where $n\in\mathbb{R}^3$ is a unit normal and $d\in\mathbb{R}$ a signed offset.
A RANSAC \cite{fischler1981random} loop proposes triples of points.

This yields a robust, metric ground plane even in cluttered scenes. It helps in robust loop detections even when robot re-visits a location from opposite ends (say approaching the same building (loop) from either it's front or back side each of them with different views, would not be an issue while perceiving the bird's eye view (BEV) \cite{li2023delving} version of the place via height maps)

\paragraph{2.\ Canonical plane-aligned frame.}
We construct an orthonormal basis
\begin{equation}
  R = [x\;\;y\;\;z] \in \mathrm{SO}(3), \qquad z := n,
\end{equation}
where the in-plane axes $x,y$ are determined by dominant eigenvectors of the projected points.
Let origin $o=\bar p$ be the mean of all inlier points.  Every world point is
expressed in plane-aligned coordinates as
\begin{equation}
  \tilde p_i = R^\top (p_i - o)
  = (u_i, v_i, h_i),
\end{equation}
where $(u_i,v_i)$ are planar coordinates and $h_i$ the signed height above
$\Pi$.  All DEM operations use $(u_i,v_i,h_i)$.

\paragraph{3.\ Rasterisation into a metric grid.}
We seek a height field
\begin{equation}
  H(u,v) : \Omega \subset \mathbb{R}^2 \to \mathbb{R},
\end{equation}
sampled on a regular grid with a \emph{globally fixed} meters-per-pixel
(\text{mpp}) resolution.

We first compute a robust planar bounding box of the projected points:
\begin{equation}
  u_0 = \min_i u_i,\quad
  u_1 = \max_i u_i,
\end{equation}
\begin{equation}
  v_0 = \min_i v_i,\quad
  v_1 = \max_i v_i.
\end{equation}
Let the longer span be
\begin{equation}
  S = \max(u_1-u_0,\; v_1-v_0),
\end{equation}
and choose a target number of pixels along this span,
$\text{target\_px\_long}\in\mathbb{N}$.  The global spatial resolution is
then
\begin{equation}
  \mathrm{mpp} = \frac{S}{\text{target\_px\_long}},
\end{equation}
so each pixel, anywhere in the DEM, corresponds to exactly $\mathrm{mpp}$
meters in the plane.

Let $W_{\text{px}}$ and $H_{\text{px}}$ be the total grid width and height in
pixels:
\begin{equation}
  W_{\text{px}} = \Bigl\lceil \frac{u_1 - u_0}{\mathrm{mpp}} \Bigr\rceil,
  \qquad
  H_{\text{px}} = \Bigl\lceil \frac{v_1 - v_0}{\mathrm{mpp}} \Bigr\rceil.
\end{equation}
We tile this grid into $N_u \times N_v$ square tiles of fixed pixel size
$\text{tile\_px}$:
\begin{equation}
  N_u = \Bigl\lceil \frac{W_{\text{px}}}{\text{tile\_px}} \Bigr\rceil,
  \qquad
  N_v = \Bigl\lceil \frac{H_{\text{px}}}{\text{tile\_px}} \Bigr\rceil.
\end{equation}
Thus each tile $(I_u,I_v)$ covers a fixed metric region of size
$\text{tile\_px}\!\times\!\text{tile\_px}$ pixels, i.e.
$(\text{tile\_px}\cdot\mathrm{mpp}) \times
 (\text{tile\_px}\cdot\mathrm{mpp})$ square meters.  The resolution
$\mathrm{mpp}$ is global and does \emph{not} change from tile to tile.

For each point $(u_i,v_i)$ we first compute its global pixel coordinates
\begin{equation}
  \hat x_i = \frac{u_i - u_0}{\mathrm{mpp}}, \qquad
  \hat y_i = \frac{v_i - v_0}{\mathrm{mpp}}.
\end{equation}
The corresponding tile indices and within-tile pixel indices are
\begin{equation}
  I_u = \Bigl\lfloor \frac{\hat x_i}{\text{tile\_px}} \Bigr\rfloor
\end{equation}
\begin{equation}
  I_v = \Bigl\lfloor \frac{\hat y_i}{\text{tile\_px}} \Bigr\rfloor,
\end{equation}

followed by clipping $x,y$ into $[0,\text{tile\_px}-1]$.  This is
the logic implemented in the rasteriser: points are binned by tile
$(I_u,I_v)$ and then by integer pixel coordinates $(x,y)$ inside each tile.

\begin{equation}
  x = \operatorname{round}\!\bigl(\hat x_i - I_u\,\text{tile\_px}\bigr),
\end{equation}
\begin{equation}
  y = \operatorname{round}\!\bigl(\hat y_i - I_v\,\text{tile\_px}\bigr),
\end{equation}

\paragraph{Height aggregation (the “reducer”).}
Multiple points may fall into the same pixel $(x,y)$ of a tile.  
Let the set of heights for that pixel be
\begin{equation}
  \{h_{x,y}^{(k)}\}_{k=1}^{K(x,y)}.
\end{equation}
DEM construction applies a \emph{reducer} function
$\mathrm{red}(\cdot)$ to obtain a single height value
\begin{equation}
  H(x,y) =
    \mathrm{red}\Bigl(\{h_{x,y}^{(k)}\}_{k=1}^{K(x,y)}\Bigr).
\end{equation}
In practice we support three choices:
\begin{itemize}
  \item \textbf{Mean reducer}:
    \begin{equation}
      \mathrm{red}_{\mathrm{mean}}
      = \frac{1}{K(x,y)} \sum_{k=1}^{K(x,y)} h_{x,y}^{(k)}.
    \end{equation}
    This yields a smooth height field but can blur sharp steps.
  \item \textbf{Max reducer}:
    \begin{equation}
      \mathrm{red}_{\mathrm{max}}
      = \max_{k} h_{x,y}^{(k)}.
    \end{equation}
    This preserves vertical discontinuities but is sensitive to outliers.
  \item \textbf{Softmax reducer} \cite{wang2023improved} (default):
    \begin{equation}
      \mathrm{red}_{\mathrm{softmax}} =
      \frac{\sum_{k} \exp\!\bigl(h^{(k)}_{x,y}/\tau\bigr)\,h^{(k)}_{x,y}}
           {\sum_{k} \exp\!\bigl(h^{(k)}_{x,y}/\tau\bigr)},
    \end{equation}
    where $\tau>0$ is the \emph{softmax temperature}.  As
    $\tau\!\to\!0$ the aggregation approaches the maximum (preserving
    sharp curbs and edges); as $\tau\!\to\!\infty$ it approaches the mean
    (smoother but more blurred).  A small but non-zero $\tau$ provides a
    good compromise: sharp road geometry with reduced sensitivity to
    spurious height spikes.
\end{itemize}

Implementation-wise, the rasteriser builds per-pixel “buckets” of heights and
applies the chosen reducer to each bucket. The raw DEM tile contains \emph{the height
field} (with NaNs (not a number) for empty pixels, independent of any visual
colourisation).  This height field is the signal used in the DINOv2-based
retrieval pipeline.

\paragraph{4.\ Post-processing and Color-Map assigned to DEMs.}
For consistent visual scaling across tiles we compute global DEM percentiles
\begin{equation}
  h_{\min} = \mathrm{perc}_{0.5}(H),\qquad
  h_{\max} = \mathrm{perc}_{99.5}(H),
\end{equation}
and normalise
\begin{equation}
  I_0(x,y) = 
    \frac{\mathrm{clip}(H(x,y),h_{\min},h_{\max}) - h_{\min}}
         {h_{\max} - h_{\min}} 
  \in [0,1].
\end{equation}

NaN pixels (no observations) are displayed as pure white.

- \textbf{Edge enhancement.}  
  Sobel gradients \cite{sobel19683x3} 
  $\nabla I_0$ produce an edge mask
  \begin{equation}
    E = 1 - \alpha_{\mathrm{edge}}\,
            \frac{\|\nabla I_0\|_2}{\mathrm{perc}_{99}(\|\nabla I_0\|_2)}.
  \end{equation}

The grayscale image $I_0$ is passed to DINOv2 \cite{oquab2023dinov2}. 

  Here $\alpha_{\mathrm{edge}}$ is the edge strength hyperparameter that determines how strongly
high-gradient regions are darkened.  Larger values produce heavier edge
shading, while $\alpha_{\mathrm{edge}}=0$ disables the effect.

 - \textbf{Hillshading} \cite{kennelly2004hillshading}.  
  From the height map $H(x,y)$ we estimate local normals and compute standard
  Lambertian shading \cite{chow2009recovering} with a virtual light direction $\ell$:
  \begin{equation}
    S(x,y)
    = \max\bigl(0,\; n_{\mathrm{surf}}(x,y)^\top \ell\bigr).
  \end{equation}
  This reveals terrain-like structure. This colored version is only used for visualisations by humans and never used by DINOv2 unlike the grayscale version which it actually interpretes.

These operations produce the yellow--green (yellow is the ground plane, darker the shade of green higher the height of the real world point) DEM visualisations as shown in Fig. \ref{fig:dem}.

\paragraph{5. Discussion.}

Ablations with the hyper-parameters discussed in this section have been shown in Table~\ref{tab:kitti_dem_ablation} of main paper on KITTI odometry \cite{geiger2013vision} sequences 00–10, reporting their respective average trajectory error in m (ATE RMSE). The default version of VGGT-SLAM++ uses softmax temperature $\tau=0.02$ and edge strength hyperparameter $\alpha_{\mathrm{edge}}$ = 0.95, with 90k pixels DEM resolution and 4096 numbers of spatial tiles. the half resolution ablation study uses 45k pixels resolution and 2048 numbers of spatial tiles while the high resolution ablation study uses 180k pixels resolution and 4096 numbers of spatial tiles (the number of smaller tiles are same in all the three cases of the default, higher and lower resolutions). No edge enhancer implies $\alpha_{\mathrm{edge}}$ = 0 and the slight enhancement uses $\alpha_{\mathrm{edge}}$ = 0.50. 

The results show that the reduction of the softmax temperature $\tau$ from 0.02 to 0.005 has lower overall ATE RMSE, as the edges are preserved, while keeping the smoothness intact at a lower $\tau$. The half resolution scenario indicates presence of a trade-off in terms of the meters per pixel (mpp) represented in the DEM. The ATE RMSE decreases from a 90k pixels  resolution (lower mpp) to a 45k  pixels (higher mpp) at half resolution choice, indicating presence of a sweet spot of resolution during the height map rendering that leads to the best results as evident from the DEM ablation study.

\subsection*{A4.\quad FAISS--HNSW as the Covisibility Retrieval Backbone}

Modern SLAM backends increasingly rely on high-dimensional embeddings
(e.g.\ DINOv2 \cite{oquab2023dinov2} features) to establish covisible submaps or long-range loop
closures.  Nearest–neighbour search \cite{kibriya2007fast} in such spaces is the core operation:
given database vectors $\{x_i\}_{i=1}^N\subset\mathbb{R}^d$ and a query
$q\in\mathbb{R}^d$, one seeks
\begin{equation}
  \operatorname*{arg\,min}_{i=1,\dots,N}\;\|q - x_i\|_2
\end{equation}
or equivalently top-$k$ neighbours under L2 \cite{mao2016multi, buhlmann2003boosting} or cosine similarity \cite{lahitani2016cosine, ye2011cosine}.

For moderate $N$ this is feasible by brute force, but for typical SLAM settings
($N$ raises to tens of thousands) \cite{wang2025depth} and queries arrive for every submap to be inserted; speed of exact search dimishes.  Hence, approximate Nearest Neighbour Search (ANNS) \cite{beis1997shape} is required.

FAISS (Facebook AI Similarity Search) \cite{douze2025faiss} is a widely used library that implements
a large family of ANNS algorithms \cite{beis1997shape}, unified under a common indexing abstraction.
It does \emph{not} learn features, but maintains a distributed service, or manage
transactions; it provides efficient, well-engineered vector indices supporting
(i) L2 distance, (ii) cosine similarity, (iii) inner product, (iv) CPU/GPU
implementations, and (v) extremely fast incremental updates.  This section
summarises the mathematical foundations  relevant to
VGGT-SLAM++, before motivating our choice of the HNSW index \cite{xiao2024enhancing}.

\paragraph{Exact vs.\ approximate search.}
Exhaustive search computes all $d$-dimensional distances,
\begin{equation}
  D_i = \|q-x_i\|_2^2,\qquad i=1,\dots,N,
\end{equation}
which costs $\mathcal{O}(Nd)$ operations per query.  This becomes prohibitive
when $N$ is large or queries arrive at video frequency.  ANNS algorithms reduce
this to \emph{sublinear} complexity (typically $\mathcal{O}(\log N)$ or
$\mathcal{O}(N^\rho)$ for $\rho<1$) by replacing the full database with a
compressed or navigable surrogate.

Approximation quality is measured by \emph{recall}:
\begin{equation}
  \mathrm{recall@}k
  = \frac{|\mathrm{ANN}(q,k)\cap \mathrm{GT}(q,k)|}{k},
\end{equation}
where $\mathrm{GT}$ denotes exact top-$k$ neighbours.  For SLAM it is essential
that recall \cite{buckland1994relationship} is high (strong covisibility cannot be missed), while latency must
remain tightly bounded.

\subsection*{A5.\quad Use of AnyLoc on DEM Images}

AnyLoc \cite{keetha2023anyloc} is a DINOv2-based \cite{oquab2023dinov2} visual place recognition system that operates on
standard images without task-specific retraining.  DEMs \cite{harithas2024findernet} encode height
which constitute coherent images with
stable local structure: edges, ridges, planar regions, and junctions appear as
distinctive textures to the ViT backbone.  Because DINOv2 features are largely
appearance-agnostic and sensitive to both geometrical and semantic cues, the same
descriptor (DINOv2) that matches natural images across viewpoints and illumination changes, also has the potential to match DEM tiles across traversal direction.

This compatibility allows us to use AnyLoc directly on the DEM domain: DEM for both tiles
and query chips are passed through the same DINOv2 encoder.  The resulting descriptors
provide robust correspondences in both indoor and outdoor trajectories, serving as candidates for a visual place recognition using retrieval technique.

\subsection*{A6.\quad  Choice of $\mathrm{Sim}(3)$ for Back-end Optimisation}

Let each VGGT-SLAM++ submap be represented by camera poses
$\{\mathbf{T}_{w\leftarrow c}^{(i)}\}_i$ and dense point maps
$\{P^{(i)}\}_i \subset \mathbb{R}^3$ as outputted by VGGT \cite{wang2025vggt}, all expressed in a common but
\emph{unknown} metric scale.  In a purely projective formulation, one would
relate two submaps via a $4\times 4$ projective warp
$\mathbf{H} \in \mathrm{SL}(4)$ \cite{hartley2003multiple},
\begin{equation}
  \mathbf{H} \in \mathrm{SL}(4)
  \;=\;
  \{\mathbf{H}\in\mathbb{R}^{4\times 4} : \det(\mathbf{H}) = 1\}
\end{equation}
which can encode non-uniform scaling, shear, and general projective skew.  In
the classical setting, this is needed because monocular reconstructions are
projectively ambiguous: points $x,x' \in \mathbb{P}^3$ satisfy
$\tilde x' \sim \mathbf{H}\tilde x$ and $\mathbf{H}$ is estimated from a
homogeneous system
$\mathbf{A}h = 0$, $h = \mathrm{vec}(\mathbf{H})$, by taking the right singular
vector of $\mathbf{A}$ associated with the smallest singular value \cite{stewart1993early}.

In VGGT--SLAM++, this level of freedom is both unnecessary and harmful. The frontend explicitly enforces
parallax: keyframes are selected only when their disparity \cite{kellnhofer2013optimizing} exceeds a threshold,
so successive submaps are linked by viewpoints with a non-trivial baseline.
Hence the regime where pure projective ambiguity is severe (extremely small baselines \cite{gallup2008variable}, near-planar scenes) gets inherently avoided. So solving for the 7 degrees of freedom is sufficiently okay for an affine solution, as projective ambiguity hardly creeps in due to the chosen setting as discussed.

By restricting the backend to $\mathrm{Sim}(3)$ \cite{greene2016multi}, we instead solve a
well-posed, over-constrained problem on a 7-dimensional Lie group \cite{loboda20237}.  The
parameter vector $\boldsymbol{\xi}\in\mathbb{R}^7$ in
$\mathrm{Sim}(3)$ directly encodes observable quantities, so the associated
Jacobian \cite{milne1986jacobian} has a small, well-understood gauge nullspace and a spectrum whose
dominant directions correspond to real geometric corrections.  Intuitively,
accumulated error over 7 meaningful degrees of freedom is far easier to
stabilise than over 15 largely redundant ones in SL(4) \cite{hartley2003multiple, donaldson1997geometry} optimisation problem. SL(4) did not converge in long KITTI and EuRoC sequences evident in the tables 1 from main paper (KITTI) and \ref{table:euroc_mav} (EuRoC). Empirically, Sim(3) based back-end, yields
substantially less drift and robust convergence on all KITTI and EuRoC
trajectories, providing both physical and numerical justification for
choosing $\mathrm{Sim}(3)$ over $\mathrm{SL}(4)$ in VGGT-SLAM++.